\documentclass[runningheads]{llncs}

\usepackage{eccv}

\usepackage{eccvabbrv}

\usepackage[utf8]{inputenc}         
\usepackage[T1]{fontenc}            
\usepackage{url}                    
\usepackage{booktabs}               
\usepackage[table,x11names,dvipsnames]{xcolor} 
\usepackage{array}
\usepackage{multirow}               
\usepackage{amsfonts}               
\usepackage{nicefrac}               
\usepackage{graphicx}               
\usepackage{enumitem}               
\usepackage{amssymb}                
\usepackage{pifont}                 
\usepackage{wrapfig}                
\usepackage{adjustbox}              
\usepackage{amsmath}
\usepackage[linesnumbered,ruled,vlined]{algorithm2e}
\usepackage{subcaption}
\usepackage{caption}
\usepackage{makecell}
\usepackage{cuted} 
\usepackage{capt-of}

\usepackage{microtype}              

\usepackage{tocloft,titletoc}       
\providecommand{\authcount}[1]{} 

\definecolor{drawioblue}{HTML}{DAE8FC}
\definecolor{mygreen}{RGB}{0,150,0}
\definecolor{myred}{RGB}{190,0,0}
\definecolor{eccvred}{rgb}{0.85,0.15,0.20}
\newcommand{\cmark}{\large\color{mygreen}\checkmark}
\newcommand{\xmark}{\large\color{myred}\ding{55}} 
\usepackage{threeparttable}

\makeatletter
\renewcommand*{\@fnsymbol}[1]{\ensuremath{\ifcase#1\or \star \or \dagger \or \ddagger \or \mathsection \or \mathparagraph \or \| \or \star\star \or \dagger\dagger \or \ddagger\ddagger \else\@ctrerr\fi}}
\makeatother

\usepackage{soul}
\setuldepth{foobar}

\usepackage[accsupp]{axessibility}  

\usepackage{hyperref}

\usepackage{orcidlink}

\begin{document}

\title{ISAC: Training-Free Instance-to-Semantic Attention Control for Multi-Instance Generation} 

\titlerunning{ISAC: Training-Free Instance-to-Semantic Attention Control}

\author{
Sanghyun Jo\inst{1,2}\thanks{Equal contribution.}\thanks{Corresponding authors: \texttt{shjo.april@gmail.com}, \texttt{kyskim@snu.ac.kr}}
\orcidlink{0000-0001-5371-2069} \and
Wooyeol Lee\inst{2}\protect\footnotemark[1]\orcidlink{0009-0005-5721-8873} \and
Ziseok Lee\inst{2}\protect\footnotemark[1]\orcidlink{0009-0005-6991-0170} \and
Jonghyun Choi\inst{2}\orcidlink{0000-0002-7934-8434} \and \\
Jaesik Park\inst{2}\orcidlink{0000-0001-5541-409X} \and
Kyungsu Kim\inst{2}\protect\footnotemark[2]\orcidlink{0000-0001-6622-6545}
}

\authorrunning{S.~Jo et al.}

\institute{OGQ, Seoul, Korea \and
Seoul National University, Seoul, Korea}

\maketitle
\begin{abstract}
  Recent open-weight text-to-image (T2I) diffusion models still struggle with multi-instance prompts, often omitting or merging instances and mixing semantics among similar objects. We trace these failures to early denoising steps, before instance boundaries are reliably stabilized.
  Existing training-free guidance is largely driven by cross-attention or other token-conditioned semantic signals. Such guidance can separate concepts at the token level, but largely assumes that distinct instance regions have already emerged. In early denoising steps, it cannot reliably carve out these regions, so count failures and semantic mixing persist. By contrast, self-attention exposes class-agnostic instance layouts during early denoising.
  To exploit this asymmetry, we propose \textbf{ISAC} (\textbf{I}nstance-to-\textbf{S}emantic \textbf{A}ttention \textbf{C}ontrol), a training-free, model-agnostic objective that first stabilizes self-attention layouts and then binds cross-attention semantics within them, without fine-tuning or external vision models.
  Across T2I-CompBench, HRS-Bench, and our newly curated IntraCompBench, ISAC consistently outperforms prior training-free methods. Furthermore, ISAC enhances layout-to-image controllers by refining coarse, overlapping bounding boxes into dense instance masks.
  Code and IntraCompBench are available at \url{https://shjo-april.github.io/ISAC}. 

\end{abstract}

\section{Introduction}

\begin{figure*}[t]
  \centering
  \includegraphics[width=\textwidth]{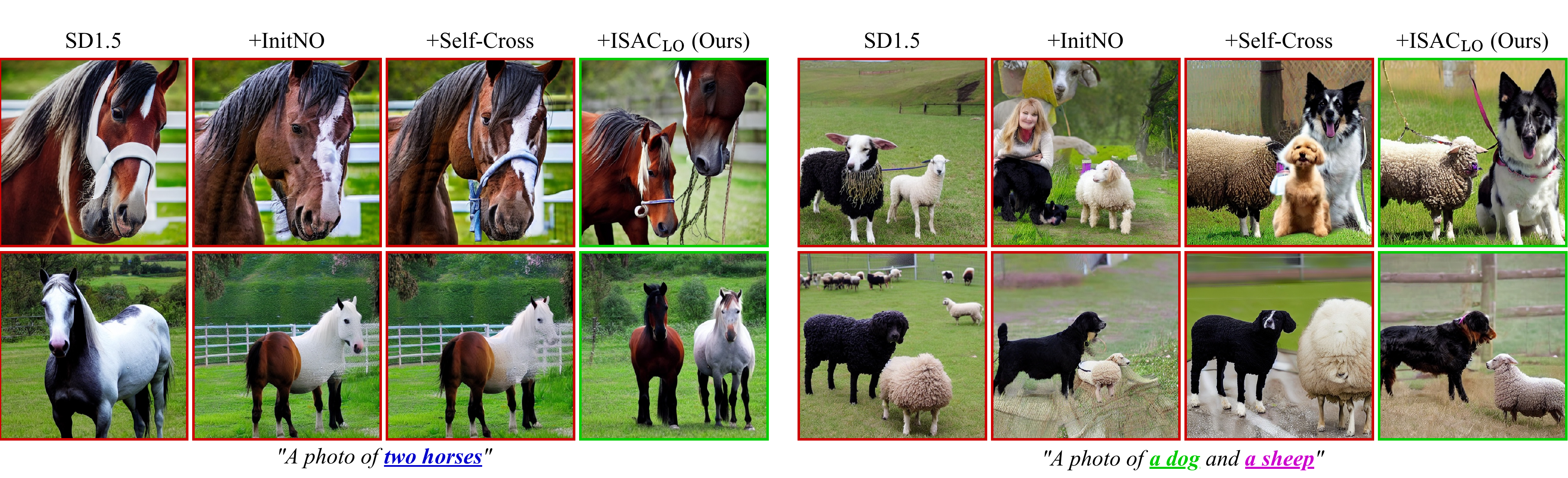}
  \caption{\textbf{Qualitative comparison for representative multi-instance prompts.} 
  Generated samples are compared across the baseline SD1.5~\cite{rombach2022high}, prior methods (InitNO~\cite{guo2024initno}, Self-Cross~\cite{qiu2025self}), and our ISAC$_\text{LO}$ (ISAC with latent optimization). Text prompts are provided below each group.
  }
  \label{fig:problem}
\end{figure*}

Text-to-image (T2I) generative models \cite{rombach2022high,flux2024,chen2025janus,nanobanana2} now produce highly realistic images from text prompts. In particular, recent commercial systems (\eg, Nano Banana 2 \cite{nanobanana2} and GPT-Image \cite{gptimage15}) establish a strong empirical upper bound on multi-instance, multi-attribute prompts. 
Such performance is often supported by massive proprietary data pipelines \cite{achiam2023gpt4,gemini25report} and expensive MLLM-driven reasoning loops, such as verifying interim images and refining the composition \cite{gemini_imagegen_api}.
Although few failure cases are observable in these systems (see Appendix \ref{sec:add-discussions}), their closed nature makes them difficult to analyze or reproduce.
This limits the community’s ability to diagnose failure modes and develop transferable solutions.

By contrast, recent open-weight diffusion backbones (\eg, Flux.1-dev \cite{flux2024} and Qwen-Image \cite{wu2025qwenimagetechnicalreport}) improve accessibility and enable reproducible inference-level analysis by releasing model weights, architectures, and inference pipelines.
Yet multi-instance compositionality remains unreliable: models often omit or merge requested objects (\emph{count failures}) and leak attributes across instances (\emph{semantic mixing}). These failures are especially pronounced when several instances are semantically similar, which is common in real-world scenarios. 
Moreover, their training data and full training recipes often remain unavailable, making retraining-based improvements hard to reproduce or transfer across backbones. This motivates training-free, inference-time guidance that improves instance separation without additional data, fine-tuning or external vision models.

Recent training-free T2I guidance methods~\cite{hu2024token,Chen_2024_TEBOpt,chefer2023attendandexcite,guo2024initno,qiu2025self} have aimed to improve compositional generation without retraining. At a high level, they mainly correct the semantic side of generation, for example by reducing interference among prompt tokens. Such correction can be effective when the model has already formed separate object regions. However, semantic correction alone does not explicitly guarantee that each requested instance is formed as a distinct region, especially for same-class or semantically similar objects.
\Cref{fig:problem} illustrates this limitation. For ``two horses,'' SD1.5 and prior semantic-driven methods often omit or merge instances. For ``a dog and a sheep,'' they often confuse the animal identities or fail to keep the requested objects spatially distinct.

\begin{figure*}[t]
    \centering
    \includegraphics[width=\linewidth]{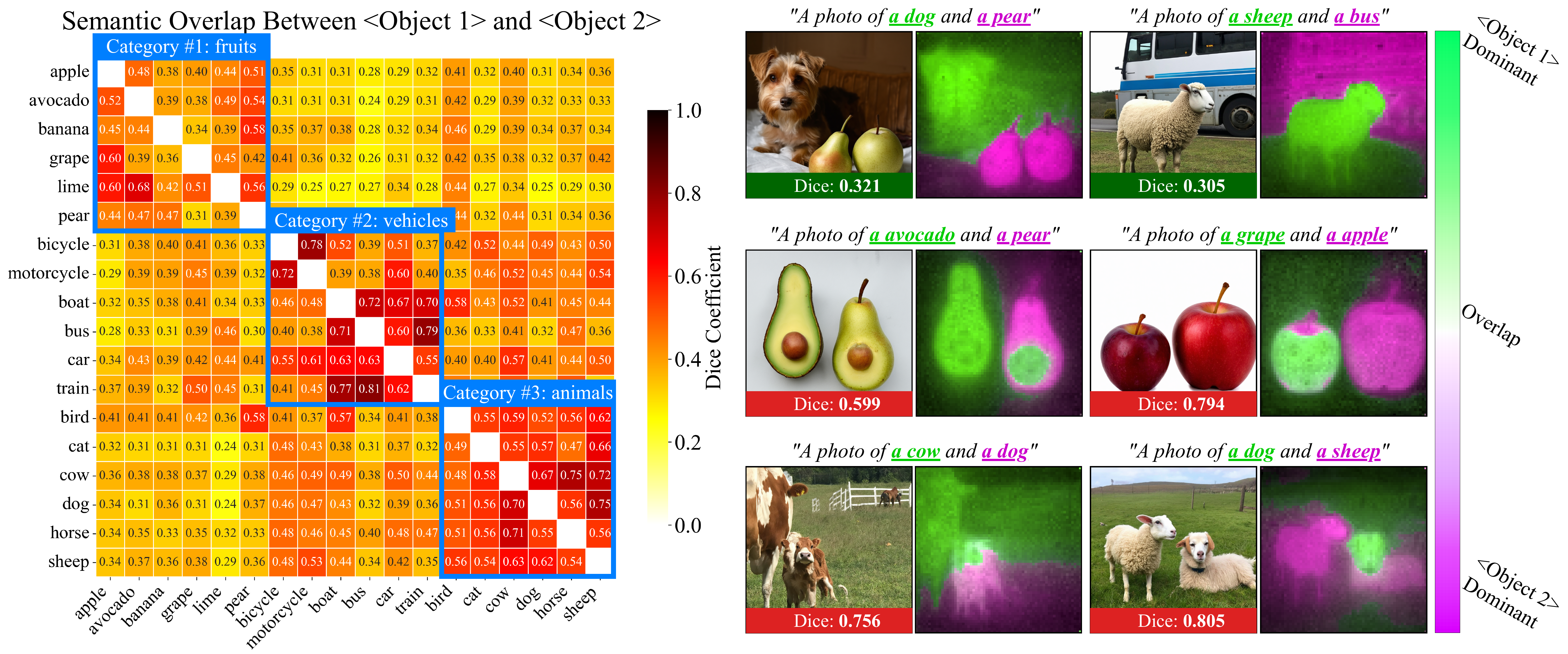}
    \caption{
    \textbf{Overlap of instance-aware semantic masks.}
    For images generated by SD3.5-M~\cite{esser2024scaling} using prompts of the format ``A photo of a \textless object1\textgreater{} and a \textless object2\textgreater{}'', instance-aware semantic masks are computed via \cref{eq:propagate} and averaged across all denoising timesteps. The Dice coefficient measures the overlap between the two object-specific masks. 
    \emph{Left:} Pairwise overlap heatmap, where blue boxes denote supercategories.
    \emph{Right:} Example generations and signed, normalized mask-difference maps, where color denotes the dominant mask and intensity indicates the normalized magnitude.
    }
    \label{fig:inter_intra_analysis}
\end{figure*}

To assess this limitation more systematically, we analyze semantic overlap across class pairs grouped by supercategory in \cref{fig:inter_intra_analysis}. We construct instance-aware semantic masks by injecting structural cues from self-attention into token-level semantics via Eq.~\ref{eq:propagate}. We then quantify semantic mixing as the Dice overlap between the two masks for a requested pair of classes. The overlap is consistently higher for pairs within the same supercategory, indicating that token-level semantic footprints tend to cover multiple semantically similar instances simultaneously. These overlaps show that token-level semantic separation may fail to determine which pixels belong to which instance when boundaries are ambiguous, as semantic cues can spill across multiple similar objects. This motivates an instance-first hierarchy that first stabilizes instance regions from structure rather than class-token semantics, and then binds semantics within each region.

We instantiate this hierarchy as ISAC (\textbf{I}nstance-to-\textbf{S}emantic \textbf{A}ttention \textbf{C}ontrol), a training-free hierarchical objective for latent optimization or latent selection. ISAC separates structural formation from semantic binding along the denoising trajectory.
Phase~1 ($\mathcal{L}_\text{ins}$) exploits the observation that self-attention already exposes emerging instance layouts in early steps~\cite{jo2026trace}, when token-conditioned semantics are still coarse and entangled. It derives \(N\) class-agnostic instance layouts from structural cues and penalizes their overlap, encouraging the requested number of separated regions before assigning class labels.
Phase~2 ($\mathcal{L}_\text{sem}$) injects these stabilized structural signals into cross-attention maps to form instance-aware semantic masks, so that classes and attributes are bound within each instance without leaking across regions.
This plug-and-play formulation supports both text-to-image backbones~\cite{rombach2022high,podell2023sdxl,esser2024scaling,chen2023pixartalpha,chen2024pixart,flux2024,flux-2-2025,arkhipkin2025kandinsky50familyfoundation,wu2025qwenimagetechnicalreport,flux2klein4b,team2025zimage} and layout-to-image controllers\cite{li2023gligen}.
To evaluate the challenging regime where these failures are most frequent, we also introduce IntraCompBench, a benchmark designed to stress-test intra-supercategory compositions.

In summary, our main contributions are:
\begin{itemize}
    \item We provide a quantitative diagnosis of semantic mixing via Dice overlap between instance-aware semantic masks, revealing substantially higher overlap for class pairs within the same supercategory.
    \item We introduce \textbf{ISAC}, a training-free, model-agnostic objective that enforces an instance-to-semantic hierarchy by separating instance formation from semantic assignment.
    \item We propose \textbf{IntraCompBench}, a benchmark for explicit 2--5-instance counting and intra-supercategory multi-class compositions.
    \item We demonstrate broad gains across T2I-CompBench~\cite{huang2025t2ipp}, HRS-Bench~\cite{bakr2023hrs}, and IntraCompBench, over multiple diffusion backbones and both text- and layout-to-image generation. ISAC improves multi-object metrics by at least $1.9\times$ over the baseline~\cite{rombach2022high} and outperforms prior counting-supervised methods~\cite{binyamin2024count, kang2025counting} by at least 10\% accuracy without additional training.
\end{itemize}

\section{Related Work}
\label{sec:related_work}

\begin{table*}[t]
\centering
\caption{
Conceptual comparison of methods for multi-instance text-to-image generation.
}
\label{tab:related_work_comparison}
\setlength{\tabcolsep}{4pt}  
\resizebox{\linewidth}{!}{
\begin{tabular}{l|ccccc}
\toprule
Method &
\makecell[c]{First preserve\\ instance structure} &
\makecell[c]{Separate\\ semantic masks} &
\makecell[c]{Diffusion-backbone\\ agnostic} &
\makecell[c]{Require instance\\ counts at inference} &
\makecell[c]{No fine-tuning or\\ extra counting model} \\
\midrule
TEBOpt {\tiny\color{gray}NeurIPS'24} \cite{Chen_2024_TEBOpt} & \xmark & \cmark & \xmark & \xmark & \cmark \\
DOS {\tiny\color{gray}AAAI'26} \cite{byun2025dos} & \xmark & \cmark & \xmark & \xmark & \cmark \\
\midrule
A\&E {\tiny\color{gray}SIGGRAPH'23} \cite{chefer2023attendandexcite} & \xmark & \cmark  & \cmark & \xmark & \cmark \\
InitNO {\tiny\color{gray}CVPR'24} \cite{guo2024initno} & \xmark & \cmark & \cmark & \xmark & \cmark \\
SynGen {\tiny\color{gray}NeurIPS'23} \cite{rassin2024linguistic} & \xmark & \cmark & \cmark & \xmark & \cmark \\
CONFORM {\tiny\color{gray}CVPR'24} \cite{meral2024conform} & \xmark & \cmark & \cmark & \xmark & \cmark \\
Self-Cross {\tiny\color{gray}CVPR'25} \cite{qiu2025self} & \xmark & \cmark & \cmark & \xmark & \cmark \\
\midrule
CountGen {\tiny\color{gray}CVPR'25} \cite{binyamin2024count} & \xmark & \cmark & \xmark & \cmark & \xmark \\
Counting Guidance {\tiny\color{gray}WACV'25} \cite{kang2025counting} & \xmark & \cmark & \xmark & \cmark & \xmark \\
\rowcolor{drawioblue} ISAC (Ours) & \cmark & \cmark & \cmark & \cmark & \cmark \\
\bottomrule
\end{tabular}
}
\end{table*}

\begin{figure*}[t]
    \centering
    \includegraphics[width=\linewidth]{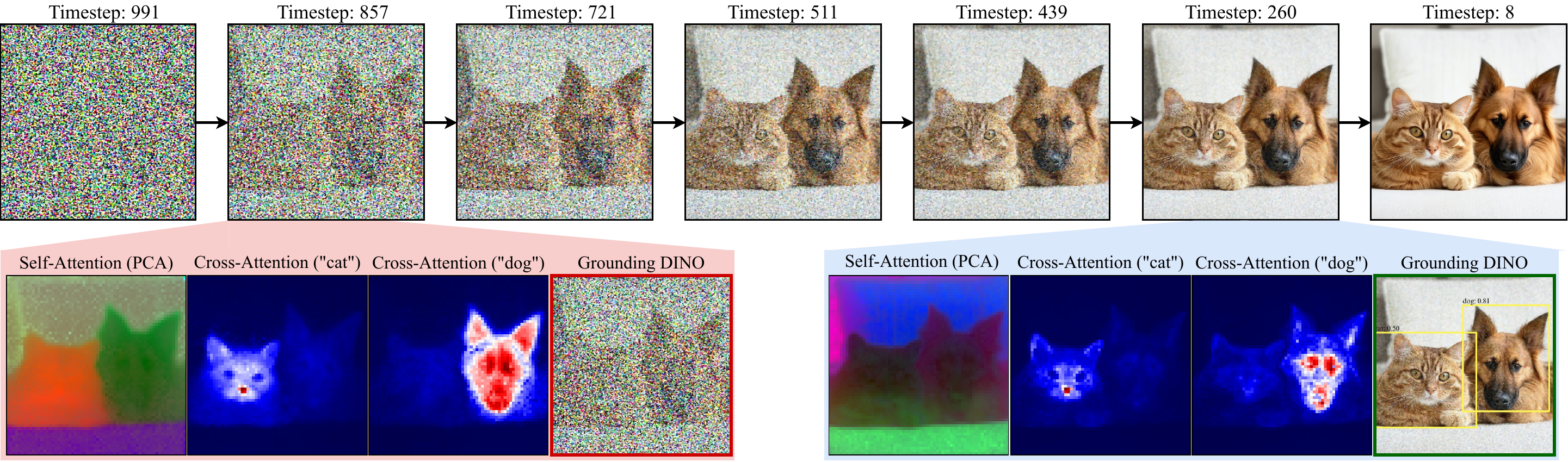}
    \caption{
    \textbf{Dynamics of text-to-image diffusion models.}
    In the early stages of diffusion, instance structures emerge \cite{jo2026trace} while semantics remain underdeveloped. In later diffusion steps, instance structures are stabilized, and semantic refinements happen. Because detection models (\eg, \cite{liu2024grounding}) rely on strong semantic cues, they are effective only in later steps.
    We use a prompt of ``A photo of a cat and a dog'' on SD3.5-M \cite{esser2024scaling}.
    }
    \label{fig:dynamics_external_model}
\end{figure*}

\subsubsection{Training-free text-to-image guidance.}
While text embedding approaches~\cite{feng2023trainingfree,hu2024token,Chen_2024_TEBOpt,byun2025dos}, such as TEBOpt \cite{Chen_2024_TEBOpt} and DOS~\cite{byun2025dos}, mitigate semantic bias via embedding optimization or token reordering, they still face two key limitations. First, without explicit spatial information, text embeddings hinder accurate instance formation. Second, reliance on the CLIP~\cite{radford2021learning} architecture limits compatibility with state-of-the-art architectures (\eg, Flux.2~\cite{flux-2-2025}, Qwen-Image~\cite{wu2025qwenimagetechnicalreport}) that leverage modern advances in large language models (LLMs).
Alternatively, cross-attention (CA) reveals the spatial footprint of textual semantics~\cite{hertz2022prompt}, enabling spatial guidance.
Attend-and-Excite~\cite{chefer2023attendandexcite} boosts attention peaks to recover neglected objects, while SynGen~\cite{rassin2024linguistic} and CONFORM~\cite{meral2024conform} separate and bind CA maps using contrastive objectives and parser-derived relations.
%
InitNO \cite{guo2024initno} and Self‑Cross~\cite{qiu2025self} additionally add structural cues from self‑attention (SA). However, these methods use CA token semantics to select, aggregate, or guide structural SA maps. 
Consequently, their structural grouping remains conditioned on CA semantics. They can separate SA maps associated with different CA tokens, but are not designed to disambiguate multiple instances that share the same semantic token. 
Thus, when early CA maps already merge same-class instances or activate only on object parts, the resulting SA-based structures inherit this ambiguity.
In contrast, ISAC is designed to explicitly establish the instance structure first and subsequently bind semantics. This strategy ensures reliable instance discrimination with minimal cues (see \cref{tab:related_work_comparison}).

\subsubsection{Layout-to-image methods.}
Token-level semantics in text prompts lack instance discriminative cues, making additional layout conditions (\eg, bounding boxes) appealing. Box layouts are lightweight and, combined with the spatial understanding of large language models (LLMs), support a two-stage pipeline that first generates a bounding-box layout from text and then conditions image synthesis on that layout~\cite{lian2023llmgrounded,zhang2024realcompo}. 
%
While state-of-the-art controllers provide dense layout control~\cite{li2023gligen,zhou2024migc,wang2024instancediffusion,cheng2024hicohierarchicalcontrollablediffusion,zhou20243dis,zhang2025creatilayout,huang2025laytrol}, they often struggle with overlapping layouts~\cite{li2025overlaybench,yang2024mastering}. 
Training-free layout-to-image methods~\cite{bar2023multidiffusion,shirakawa2024noisecollage,Xie_2023_ICCV,chen2023trainingfree,xiao2023rb,lee2024groundit,phung2024grounded,dahary2024yourself,kim2025improving} help, yet they do not ensure instance discrimination. 
Constraining attention within each box either by excluding background~\cite{phung2024grounded} or other instances’ layouts~\cite{dahary2024yourself} only separates semantic regions rather than instance structures. 
In contrast, ISAC first evaluates instance structures from internal attention without layout priors, then assigns semantics. This addresses controllers’ failures under overlapping boxes by carving coarse box layouts into dense instance masks.

\subsubsection{Count-supervised text-to-image methods.}
Given the limits of box layouts, recent work pursues instance-level control with minimal supervision via instance counts. One approach leverages a pretrained vision model to enforce counts~\cite{kang2025counting}, but because such models rely on strong semantic cues, they are ineffective in early diffusion steps when instances form (see~\cref{fig:dynamics_external_model}).
Another replaces boxes with automatically generated dense masks to guide instance separation~\cite{binyamin2024count,dahary2025decisive}, yet this depends on a fine-tuned mask generator and extensive training to cover broad vocabularies and compositions.

\subsubsection{Evaluation and benchmarks.}
To evaluate multi-object generation, T2I-CompBench~\cite{huang2025t2ipp} and HRS-Bench~\cite{bakr2023hrs} are widely adopted. However, they do not isolate intra-category cases where (i) \emph{count failures} and (ii) \emph{semantic mixing} are most prevalent. 
Evaluation set from~\cite{chefer2023attendandexcite} and SSD~\cite{qiu2025self} target this issue but benchmark only simple 2--3-object compositions, with SSD limited to a small prompt set (31 two-object and 21 three-object prompts). 
We address this gap with IntraCompBench, a comprehensive benchmark that stress-tests similar-object generation across 2--5-object compositions with diverse prompt combinations.

\section{Method}
\label{sec:method}

\begin{figure*}[t]
    \centering
    \includegraphics[width=\linewidth]{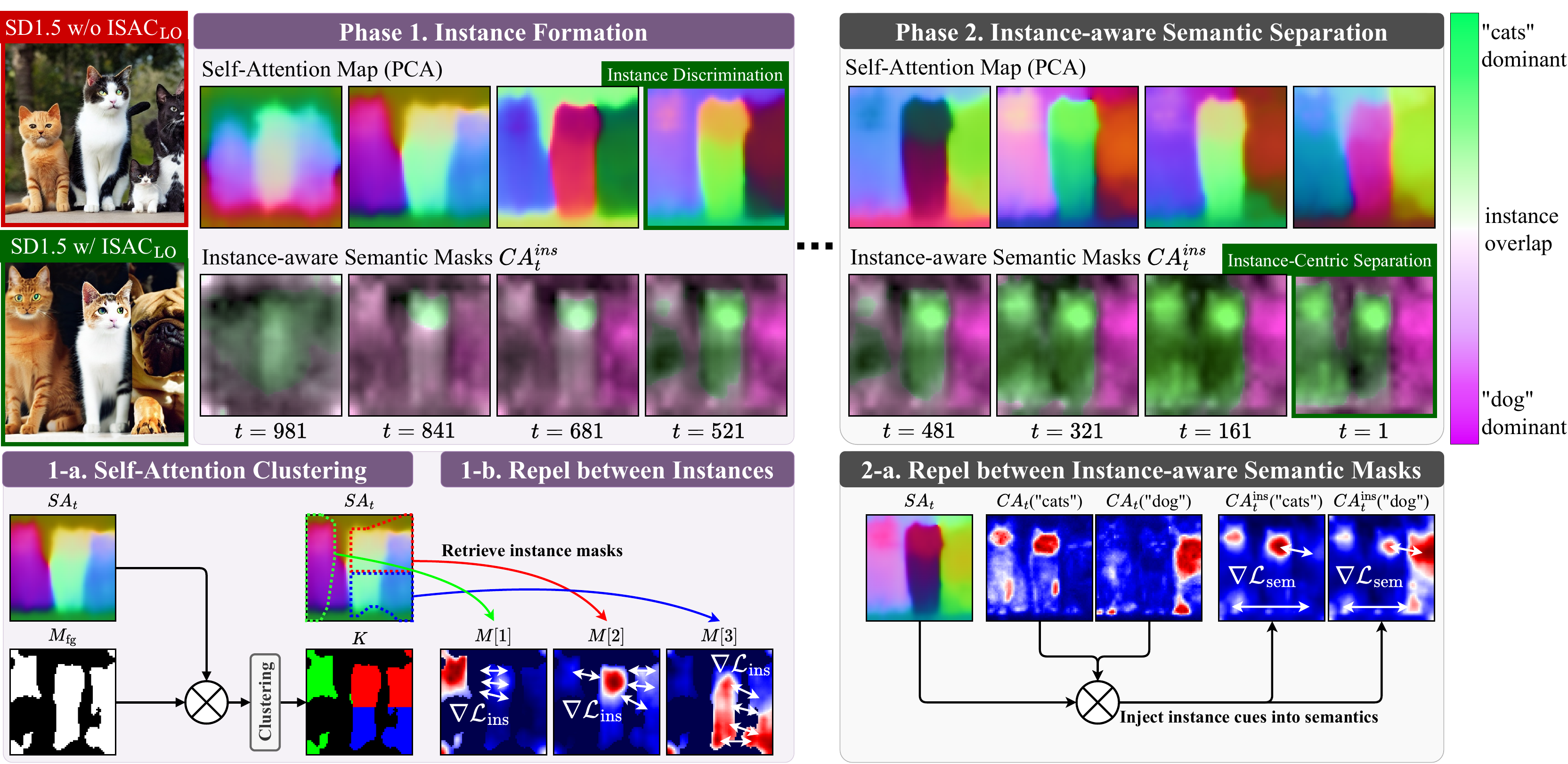}
    \caption{
    \textbf{Overview of ISAC.} 
    Guided by diffusion dynamics, ISAC computes a hierarchical objective in two phases. Phase~1 (\cref{sec:phase_1}) clusters self-attention to shape \(N\) class-agnostic instance layouts, repelling overlaps to establish clean boundaries early in the trajectory.
    Phase~2 (\cref{sec:phase_2}) then inject these reliable instance structures into cross-attention to align semantic evidence, using a repel-and-bind loss to prevent cross-instance semantic mixing. An instance-to-semantic schedule (\cref{sec:isac_loss_schedule}) seamlessly transitions the objective from Phase~1 to Phase~2.
    }
    \label{fig:method-overview}
\end{figure*}

ISAC targets two core failure modes of multi-object generation, \emph{count failures} and \emph{semantic mixing}, which stem from vague instance boundaries. Guided by evidence that coarse structure emerges before fine-grained semantics in diffusion \cite{hertz2022prompt,lee2025beta}, we design ISAC as a hierarchical, instance-first, training-free objective function. \Cref{fig:method-overview} illustrates how Phase~1 (\cref{sec:phase_1}) shapes class-agnostic instance layouts from self-attention and how Phase~2 (\cref{sec:phase_2}) aligns semantic evidence to these layouts, yielding instance-aware semantics that prevent cross-instance mixing.

\subsection{Background}
\label{sec:preliminaries}

\subsubsection{Text-to-image diffusion models.}
Diffusion models learn to reverse a forward noising process that gradually corrupts a sample \(X_0\) into noise \(X_T\), conditioned on a text embedding \(\mathcal T \in \mathbb R^{L\times d}\).
At inference time, a neural network \(\epsilon_\theta\) iteratively denoises \(X_T \sim \mathcal N(0, I)\) to obtain \(X_0\).
In text-to-image models, denoising is typically performed in latent space; a VAE decoder \(\mathcal D\) then maps \(X_0\) to pixel space, yielding the final image \(I_0 = \mathcal D(X_0)\).
ISAC is model-agnostic: it only requires access to \(X_t\) and the model’s attention maps.

\subsubsection{Prompt notation.}
Following \cite{kang2025counting,binyamin2024count}, ISAC targets the multi-instance setting in which each prompt explicitly specifies instance counts. For each prompt, we parse class tokens \(\{\tau_i\}_{i=1}^k\), per-class instance counts \(\{n_i\}_{i=1}^k\), and optional attributes \(\{\chi_{i,j}\}\). Ambiguous-count prompts, where the requested number of instances cannot be determined from the prompt, are outside the scope of this work. The total number of requested instances is then given by \(N=\sum_i n_i\). We use an LLM-based parser to automatically obtain \((\{\tau_i\}, \{n_i\}, \{\chi_{i,j}\})\). Additional details and examples are provided in Appendix~\ref{sec:implementation_details}.

\subsubsection{Attention.}
At each denoising timestep $t$, the denoiser \(\epsilon_\theta\) uses {self‑attention} (SA) to capture spatial relations within the latent \(X_t \in \mathbb R^{HW\times d}\) and {cross‑attention} (CA) to align \(X_t\) with text embeddings \(\mathcal T\).
Both compute maps from queries and keys obtained via learned projections \((W_Q, W_K)\).
For SA, \(Q_t^{\text{self}} = X_t W_Q^{\text{self}}\) and \(K_t^{\text{self}} = X_t W_K^{\text{self}}\).
For CA, \(Q_t^{\text{cross}} = X_t W_Q^{\text{cross}}\) and \(K_t^{\text{cross}} = \mathcal T W_K^{\text{cross}}\).
For a head of width \(d_h\) at layer $l$, the per‑head attention maps are
\begin{flalign}
    SA^h_l(X_t) &= \mathrm{softmax} ( {Q_t^{\text{self}} {K_t^{\text{self}}}^\top} /\sqrt{d_h}) \in [0,1]^{HW \times HW}, \\
    CA^h_l(X_t,\mathcal{T}) &= \mathrm{softmax} ({Q_t^{\text{cross}} {K_t^{\text{cross}}}^\top} / \sqrt{d_h}) \in [0,1]^{HW \times L}.
\end{flalign}

\noindent We register hooks on all attention layers to read out SA and CA without altering computation.
Let \(M\) be the number of attention layers and \(h_l\) the number of heads at layer \(l\). For each timestep $t$, we obtain a single SA/CA pair by averaging the attention maps over all layers and heads: 
\begin{align}
    SA_t\!=\! \frac{1}{\mathcal N}\! \sum_{l,h}SA_l^h(X_t),\; CA_t\!=\! \frac{1}{\mathcal N}\! \sum_{l,h} CA_l^h(X_t,\mathcal{T})
\label{eq:SA_t_and_CA_t}
\end{align}
where $\mathcal N=\sum_{l=1}^Mh_l$. For U-Nets where attention maps vary in spatial resolution, we bilinearly upsample each map to the highest resolution and take the average (See Appendix \ref{sec:implementation_details}).

\subsection{Phase 1: Instance Formation}
\label{sec:phase_1}

Motivated by \cite{jo2026trace}, we observe that pixels belonging to the same instance exhibit higher mutual attention, whereas pixels from different instances attend to each other less. This suggests that the \(N\) most discriminative pixel clusters reveal \(N\) disjoint instance layouts. We translate this property into a guidance objective by (i) clustering self-attention into \(N\) groups and (ii) penalizing overlap between the resulting masks. This loss strengthens attention within each instance and suppresses attention outside its boundary.

\subsubsection{Self-attention to $N$ clusters.}
Self-attention (SA) encodes instance structure, but also assigns non-trivial mass to background regions, so clustering it over the whole image can split or merge instances. We therefore first build a foreground gate $M_{\mathrm{fg}}$ from semantic maps and then cluster SA only within this foreground.
Given accumulated maps \(SA_t \) and \(CA_t \) from Eq. \eqref{eq:SA_t_and_CA_t}, we leverage the instance structure encoded in \(SA_t\) to form instance-aware semantic masks:
\begin{equation}
    \label{eq:propagate}
    CA_t^{\text{ins}} \;=\; SA_t\,CA_t \;\in\; [0,1]^{HW\times L},
\end{equation}
%
where column $j$ highlights the region most responsive to token $\mathcal T[j]$.  
We binarize each column by its mean \(\mu_j\) and define $M_{\mathrm{fg}}$ as the union over class tokens,
\begin{align}
    CA_t^{\text{bin}} &\gets \texttt{Binarize}(CA_t^{\text{ins}}), \label{eq:binarize} \\
    M_{\mathrm{fg}} &= \bigcup_{\mathcal{T}[i]\in\{\tau_j\}_{j=1}^k} CA_t^{\text{bin}}[:,i] \in \{0,1\}^{HW}. \label{eq:fg_mask}
\end{align}

\noindent Let $\mathcal I{=}\{p: M_{\mathrm{fg}}[p]{=}1\}$ and $F{:=}|\mathcal I|$. 
We restrict SA to these foreground positions and cluster its rows into \(N\) components (\eg, K-means on SA features, concatenated with normalized coordinates \((x,y)\in[-1,1]^2\) for spatial coherence). With one-hot assignments \(K \in \{0,1\}^{F\times N}\) (no gradient through \(K\)), the resulting instance masks are
\begin{equation}
\label{eq:instance_mask}
    M \;=\; SA_t[\mathcal I,\mathcal I]\;\texttt{stopgrad}(K) \;\in\; [0,1]^{F\times N}.
\end{equation}
This yields instance masks that highlight pixels attending strongly within the same cluster and weakly outside, leading to sharper instance boundaries. Further details of the instance clustering procedure are provided in Appendix~\ref{sec:implementation_details}.

\subsubsection{Repel guidance between instance masks.}
From \cref{eq:instance_mask} we obtain instance masks \(M[1],\dots,M[N]\) over foreground pixels. 
To separate instances, we penalize their worst local overlap using the \emph{maximum pixel-wise overlap} (MPO):
\begin{equation}
\label{eq:mpo}
    \texttt{MPO}(A,B) \;=\; \max_{p \in \{1,\dots,F\}} A[p]\cdot B[p],
\end{equation}
and define the instance separation loss as the maximum MPO over all mask pairs,
\begin{equation}
\label{eq:L_ins}
    \mathcal L_{\mathrm{ins}}(X_t) \;=\; \max_{1 \le i < j \le N} \texttt{MPO}\big(M[i],\,M[j]\big).
\end{equation}
Within each step we treat \(K\) as \texttt{stopgrad}, so gradients flow only through \(SA_t\).

\subsection{Phase 2: Instance-aware Semantic Separation} \label{sec:phase_2}

After Phase~1 stabilizes sharp instance boundaries in $SA_t$, the propagated semantic maps $CA_t^{\text{ins}} = SA_t\,CA_t$ in \cref{eq:propagate} are activated within spatially partitioned instances. Building on this structure, in Phase~2, we apply a repel-and-bind loss $\mathcal{L}_{\text{sem}}$: tokens referring to different instances are pushed apart, while tokens describing the same instance are pulled together, reinforcing clean semantic separation per instance.

Let \(P_{\text{repel}}\) denote pairs of token indices that should remain distinct (\emph{e.g.,}, different classes/instances), and \(P_{\text{bind}}\) denote pairs that should co-activate (\eg, class/attribute within the same instance). We use MPO as a sharp, local measure of overlap between the corresponding semantic maps:
\begin{align}
\mathcal L_{\text{repel}}(X_t)\!&=\!\max_{(a,b)\in P_{\text{repel}}}\!
    \ \big[+\texttt{MPO}\big(CA_t^{\text{ins}}[:,a],\,CA_t^{\text{ins}}[:,b]\big) \big] \\
\mathcal L_{\text{bind}}(X_t)\!&=\!\max_{(a,b)\in P_{\text{bind}}}\!
    \ \big[1\!-\!\texttt{MPO}\big(CA_t^{\text{ins}}[:,a],\,CA_t^{\text{ins}}[:,b]\big)\big]
\end{align}
We combine them as a repel-and-bind objective:
\begin{equation}
\label{eq:L_sem}
\mathcal L_{\text{sem}}(X_t)\;=\;\mathcal L_{\text{repel}}(X_t) + \mathcal L_{\text{bind}}(X_t)
\end{equation}
While using such token relations is common~\cite{rassin2024linguistic,meral2024conform}, our contribution lies in binding these relations to the \emph{instance-aware} masks formed in Phase~1 (\cref{sec:phase_1}).

\begin{algorithm}[b]
\caption{ISAC with Latent Optimization (ISAC$_\text{LO}$)}
\label{alg:isac_optimization}
\DontPrintSemicolon

\KwIn{Prompt $\mathcal{T}$, Model $\epsilon_\theta$, decoder $\mathcal{D}$, step size $\eta$}
\KwOut{Image $I_0$}
$X_T \sim \mathcal{N}(0,I)$ \;

\For{$t=T,T{-}1,\ldots,1$}{
  $\texttt{Call}~\text{Denoise}(X_t,\mathcal T,\epsilon_\theta, t)
  ~\texttt{with Hooks} \to SA_t, CA_t$\;
  $CA_t^{\text{ins}} \gets SA_t \cdot CA_t$ \;
  \texttt{Build} foreground gate and instance masks (Eqs. \ref{eq:binarize}, \ref{eq:fg_mask}, \ref{eq:instance_mask}) \;
  \texttt{Compute }$\mathcal{L}_\text{ins},\mathcal{L}_\text{sem} $ (Eqs. \ref{eq:L_ins}, \ref{eq:L_sem}) \;
  $\mathcal{L}_\text{ISAC}(X_t, t) \gets \lambda_\text{ins}(t)\mathcal{L}_\text{ins}(X_t) + \lambda_\text{sem}(t)\mathcal{L}_\text{sem}(X_t)$\;
  $\tilde{X}_t \gets X_t - \eta \cdot \nabla_{X_t} \mathcal{L}_\text{ISAC}(X_t, t)$\;
  $X_{t-1} \gets \text{Denoise}(\tilde{X}_t, \mathcal T,\epsilon_\theta, t)$
}
$I_0 \gets \mathcal{D}(X_0)$ \tcp*{Decode to pixel space}
\end{algorithm}

\subsection{Instance-to-Semantic Loss Schedule}
\label{sec:isac_loss_schedule}

Combining both phases, we define the per-step ISAC objective as
\begin{equation}
\label{eq:ISAC}
    \mathcal L_\text{ISAC}(X_t,t)\;=\;\lambda_{\text{ins}}(t)\,\mathcal L_{\text{ins}}(X_t)\;+\;\lambda_{\text{sem}}(t)\,\mathcal L_{\text{sem}}(X_t),
\end{equation}
where \(\lambda_{\text{ins}}(t)\) and \(\lambda_{\text{sem}}(t)\) control the relative emphasis on instance layout versus semantics over the diffusion trajectory; in practice, we simply set \(\lambda_{\text{ins}}(t) = t/T\) and \(\lambda_{\text{sem}}(t) = 1 - t/T\) so that early steps focus on instance formation and later steps focus on semantic refinement.

Following prior work \cite{chefer2023attendandexcite,rassin2024linguistic,meral2024conform,qiu2025self}, we primarily use ISAC for latent optimization (ISAC$_\text{LO}$) as summarized in Algorithm~\ref{alg:isac_optimization}. ISAC is also compatible with other guidance schemes, such as latent selection (ISAC$_\text{LS}$; see \cref{subsubsec:latent-selection}).

\section{Experiments}
\label{sec:experiments}

\subsection{Experimental Setup}
\label{subsec:setup}

\subsubsection{Benchmarks.} We evaluate ISAC on three benchmarks that cover complementary aspects of multi-instance generation: T2I-CompBench \cite{huang2025t2ipp}, HRS-Bench \cite{bakr2023hrs}, and our IntraCompBench. T2I-CompBench \cite{huang2025t2ipp} and HRS-Bench \cite{bakr2023hrs} are standard for compositional text-to-image evaluation but do not isolate the intra-category regime where \emph{count failures} and \emph{semantic mixing} are most severe, so we introduce IntraCompBench to explicitly target two failure modes: multi-instance accuracy (prompts such as ``three dogs''), which asks the model to generate a specified count $N$ of a single class and measures how often the predicted instance count matches $N$, and multi-class accuracy (prompts such as ``a dog, a cat, and a horse''), which asks for one instance of each of $k$ different classes ($N = k$) and measures how often all $k$ classes appear as distinct instances. For both settings, we evaluate generated images using open-vocabulary detection models \cite{liu2024grounding, wang2025yoloerealtimeseeing} and a prompt-specific matching procedure between detections and requested classes (see Appendix~\ref{sec:implementation_details} for details of our IntraCompBench).

\subsubsection{Implementation details.} We strictly follow the official inference configurations for all diffusion models~\cite{rombach2022high, podell2023sdxl, esser2024scaling, flux2024, wu2025qwenimagetechnicalreport}. We apply the ISAC objective (Eq. \eqref{eq:ISAC}) in latent optimization and latent selection, two common training-free approaches for multi-instance generation. Both variants share identical schedules $\lambda_\text{ins}(t)$ and $\lambda_\text{sem}(t)$, which are fixed by design as in \cref{eq:ISAC} and are not tuned per model or benchmark. ISAC with latent optimization (ISAC$_\text{LO}$) requires a single tuned hyperparameter, $\eta = 0.01$, shared across all models and benchmarks. Sensitivity analysis for $\eta$ is provided in Appendix~\ref{sec:implementation_details}. For all experiments involving ISAC with latent selection (ISAC$_\text{LS}$), we choose the best-out-of-10. For each text prompt, we automatically extract class tag–count pairs $(\tau_i, n_i)$ and simple token relations ($P_{\text{repel}}$ and $P_{\text{bind}}$) using the open-sourced GPT-OSS~\cite{openai2025gptoss120bgptoss20bmodel} to ensure reproducibility. More details (\eg, GPT instructions) are described in Appendix~\ref{sec:implementation_details}.

\begin{table*}[tp]
\centering

\caption{
\textbf{Quantitative results on HRS-Bench~\cite{bakr2023hrs}, T2I-CompBench~\cite{huang2025t2ipp}, and IntraCompBench.} ``Class'' denotes the multi-class subset of IntraCompBench. Efficiency metrics are averaged across these multi-class task runs.
}
\label{tab:main-results}

\newcolumntype{C}[1]{>{\centering\arraybackslash}p{#1}}

\newlength{\mycolwidth}
\setlength{\mycolwidth}{0.098\textwidth} 
    
\resizebox{\linewidth}{!}{%
\begin{tabular}{
  l  
  *{3}{C{0.75\mycolwidth}}  
  *{3}{C{1.05\mycolwidth}}  
  *{5}{C{0.7\mycolwidth}}  
  c  
  c  
}
    \toprule
    \multirow{2}{*}{Method}
      & \multicolumn{3}{c}{HRSBench(\(\uparrow\))}
      & \multicolumn{3}{c}{T2I-CompBench(\(\uparrow\))} 
      & \multicolumn{5}{c}{IntraCompBench(Class)(\(\uparrow\))} 
      & \multicolumn{2}{c}{Efficiency(\(\downarrow\))} \\
    \cmidrule(lr){2-4} \cmidrule(lr){5-7} \cmidrule(lr){8-12} \cmidrule(lr){13-14}
      & Color
      & Spatial
      & Size 
      & Color 
      & Texture 
      & Complex 
      & \#2 & \#3 & \#4 & \#5 & Avg. 
      & Latency & VRAM \\
    \midrule
    SD1.5 \cite{rombach2022high}                & 0.136 & 0.094 & 0.091 & 0.356 & 0.406 & 0.306 & 28\% & 2\%  & 1\%  & 0\%  & 8\%  & \textbf{8s}  & \textbf{4.9 GB} \\
    + A\&E {\tiny\color{gray}SIGGRAPH'23} \cite{chefer2023attendandexcite}       & 0.149 & 0.104 & 0.101 & 0.392 & 0.447 & 0.290 & 48\% & 10\% & 5\%  & 2\%  & 16\% & 17s & 9.2 GB \\
    + SynGen {\tiny\color{gray}NeurIPS'23} \cite{rassin2024linguistic}          & 0.159 & 0.111 & 0.107 & 0.420 & 0.479 & 0.311 & 50\% & 9\%  & 4\%  & 2\%  & 16\% & 19s & 9.3 GB \\
    + InitNO {\tiny\color{gray}CVPR'24} \cite{guo2024initno}                 & 0.175 & 0.120 & 0.116 & 0.456 & 0.520 & 0.338 & 55\% & 12\% & 7\%  & 5\%  & 20\% & 20s & 9.6 GB \\
    + TEBOpt {\tiny\color{gray}NeurIPS'24} \cite{Chen_2024_TEBOpt}              & 0.181 & 0.127 & 0.123 & 0.461 & 0.544 & 0.353 & 52\% & 11\% & 8\%  & 3\%  & 18\% & 10s & 6.4 GB \\      
    + Self-Cross {\tiny\color{gray}CVPR'25} \cite{qiu2025self}              & 0.170 & 0.118 & 0.114 & 0.445 & 0.508 & 0.324 & 48\% & 8\% & 4\% & 2\% & 15\% & 21s & 10 GB \\
    + DOS {\tiny\color{gray}AAAI'26} \cite{byun2025dos}            & 0.191 & 0.135 & 0.121 & 0.468 & 0.531 & 0.341 & 56\% & 14\% & 7\% & 4\% & 20\% & 11s & 6.5GB \\
    \rowcolor{drawioblue}
    + \textbf{ISAC$_\text{LO}$ (Ours)}                       & \textbf{0.318} & \textbf{0.263} & \textbf{0.252} & \textbf{0.683} & \textbf{0.631} & \textbf{0.354} & \textbf{65\%} & \textbf{31\%} & \textbf{29\%} & \textbf{18\%} & \textbf{36\%} & 21s & 9.7 GB \\

    \midrule
    SD3.5-M \cite{esser2024scaling}             & 0.425 & 0.264 & 0.209 & 0.796 & 0.726 & 0.377 & 62\% & 23\% & 12\% & 3\%  & 25\% & \textbf{40s} & \textbf{22.9 GB} \\
    + A\&E {\tiny\color{gray}SIGGRAPH'23} \cite{chefer2023attendandexcite}     & 0.427 & 0.263 & 0.215 & 0.798 & 0.726 & 0.378 & 65\% & 29\% & 16\% & 5\%  & 28\%  & 124s & 73.8 GB \\
    + SynGen {\tiny\color{gray}NeurIPS'23} \cite{rassin2024linguistic}         & 0.425 & 0.260 & 0.211 & 0.801 & 0.718 & 0.365 & 66\% & 28\% & 15\% & 6\%  & 28\%  & 131s & 74.3 GB \\
    + InitNO {\tiny\color{gray}CVPR'24} \cite{guo2024initno}               & 0.443 & 0.275 & 0.228 & 0.810 & 0.728 & 0.378 & 77\% & 31\% & 17\% & 7\%  & 33\%  & 138s & 74.6 GB \\
    + TEBOpt {\tiny\color{gray}NeurIPS'24} \cite{Chen_2024_TEBOpt}            & 0.438 & 0.279 & 0.220 & 0.805 & 0.730 & 0.381 & 78\% & 31\% & 19\% & 8\%  & 34\%  & 42s & 28.8 GB \\
    + Self-Cross {\tiny\color{gray}CVPR'25} \cite{qiu2025self}             & 0.431 & 0.268 & 0.219 & 0.795 & 0.720 & 0.371 & 78\% & 38\% & 19\% & 3\% & 34\% & 147s & 76.4 GB \\
    + DOS {\tiny\color{gray}AAAI'26} \cite{byun2025dos}            & 0.440 & 0.273 & 0.231 & 0.808 & 0.729 & 0.380 & 79\% & 33\% & 18\% & 7\% & 34\% & 44s & 29.1GB \\
    \rowcolor{drawioblue}
    + \textbf{ISAC$_\text{LO}$ (Ours)}                       & \textbf{0.473} & \textbf{0.350} & \textbf{0.258} & \textbf{0.838} & \textbf{0.739} & \textbf{0.388} & \textbf{98\%} & \textbf{51\%} & \textbf{40\%} & \textbf{20\%} & \textbf{52\%} & 140s & 74.8 GB \\

    \bottomrule
\end{tabular}
}
\end{table*}

\begin{figure*}[htp]
    \centering
    \includegraphics[width=\linewidth]{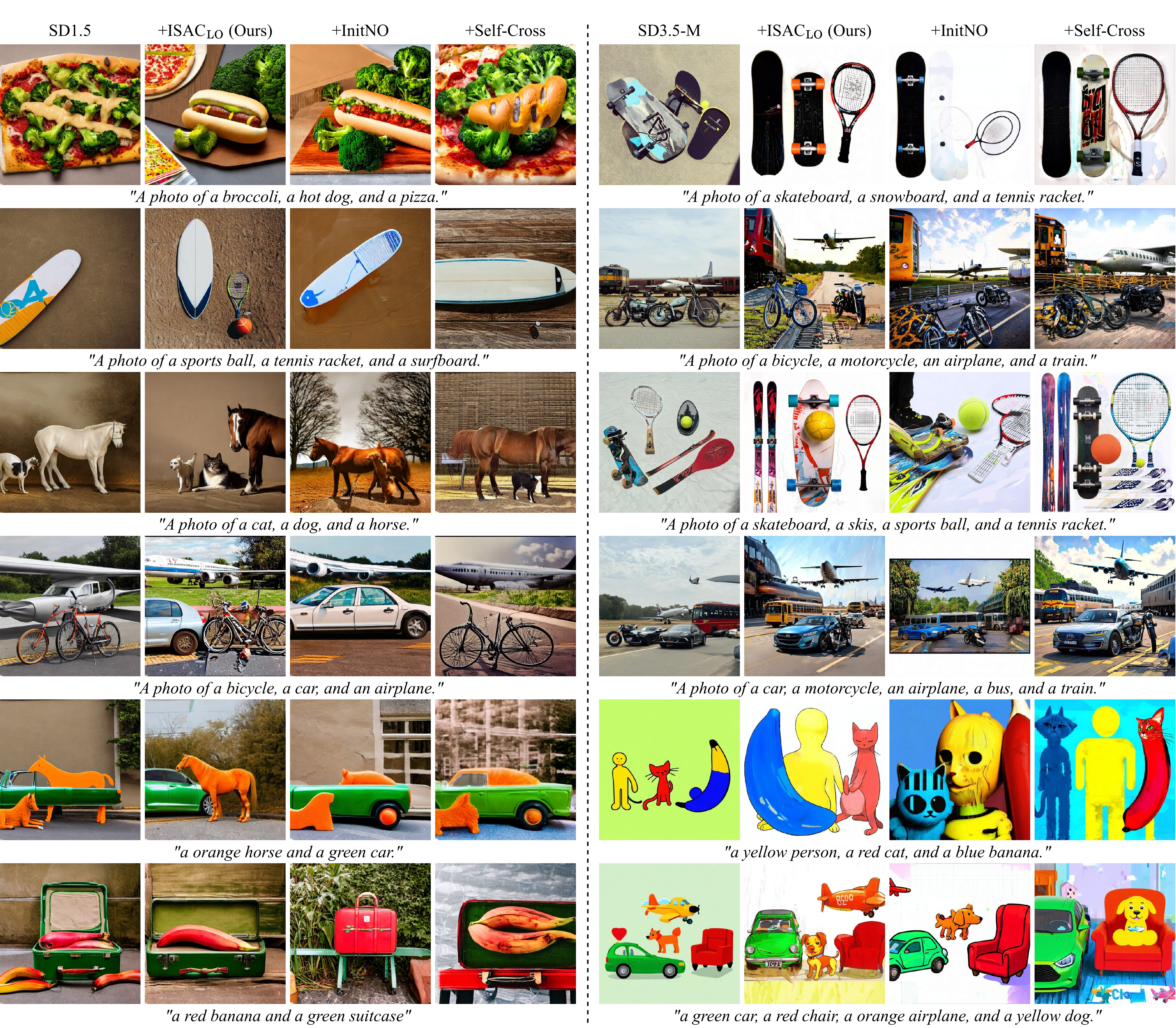}
    \caption{
    Qualitative comparison of attention control methods on SD1.5~\cite{rombach2022high} and SD3.5-M~\cite{esser2024scaling}.
    }
    \label{fig:qualitative-results-baselines}
\end{figure*}

\subsection{Comparison with State of the Arts}
\label{subsec:main_results}
Table \ref{tab:main-results} summarizes quantitative results for two popular diffusion models \cite{rombach2022high, esser2024scaling}. With similar inference cost compared to attention control methods \cite{chefer2023attendandexcite,rassin2024linguistic,guo2024initno,qiu2025self}, ISAC$_\text{LO}$ clearly outperforms them on every metric of HRS-Bench, T2I-CompBench, and the multi-class setting of IntraCompBench. Text embedding approaches \cite{Chen_2024_TEBOpt,byun2025dos} show lower computational overhead, yet their performance falls behind ISAC due to spatially unaware optimization.
The largest improvements appear in the intra-category regime: on SD1.5 \cite{rombach2022high}, the average multi-class accuracy on IntraCompBench increases from 20\% for the strongest baseline \cite{guo2024initno,byun2025dos} to 36\% with ISAC, and the HRS-Bench spatial score is roughly doubled (0.135 to 0.263). On the stronger SD3.5-M backbone, ISAC still provides a clear margin with especially large gains in the more crowded \#4 and \#5 cases. These trends indicate that our instance-first attention control is most beneficial exactly where multiple similar objects must be separated and counted.

As shown in \cref{fig:qualitative-results-baselines}, previous state-of-the-art methods \cite{guo2024initno, qiu2025self} frequently blur boundaries between objects or merge categories into hybrid shapes when several related objects appear in a scene. By contrast, ISAC allocates distinct, spatially coherent instances to each requested class while maintaining their attributes. This behavior is consistent across prompts (from simple color–shape compositions to scenes with multiple objects) and across two diffusion models \cite{rombach2022high, esser2024scaling}. Additional quantitative and qualitative results are provided in Appendices \ref{sec:add-quantitative-results} and \ref{appendix_qualitative}.

\subsection{Discussion}
\label{subsec:discussion}

\begin{table*}[t]
\centering
\setlength{\tabcolsep}{2pt}       

\begin{minipage}[t]{0.5\linewidth}
\centering

\captionof{table}{
Comparison of ISAC$_\text{LO}$ and count-supervised methods~\cite{kang2025counting,binyamin2024count} on IntraCompBench.
}
\label{tab:count_results}

\newcolumntype{C}[1]{>{\centering\arraybackslash}p{#1}}

\newlength{\countquantcolwidth}
\setlength{\countquantcolwidth}{0.13\textwidth} 

\resizebox{1.0\linewidth}{!}{
\begin{tabular}{
    l   
    *{5}{C{\countquantcolwidth}}  
    c   
    c   
}
    \toprule
    \multirow{2}{*}{Method}
      & \multicolumn{5}{c}{IntraCompBench(Instance)(\(\uparrow\))}
      & \multicolumn{2}{c}{Efficiency (\(\downarrow\))} \\
    \cmidrule(lr){2-6} \cmidrule(lr){7-8}
      & \#2 & \#3 & \#4 & \#5 & Avg.
      & Latency & VRAM \\
    \midrule
      SD1.4 \cite{rombach2022high} & {94\%} & {74\%} & 28\% & {22\%} & {55\%} & \textbf{8s} & \textbf{4.9GB} \\
    + CG {\tiny\color{gray}WACV'25}\cite{kang2025counting}  & 79\% & 67\% & {32\%} & 19\% & 49\% & {14s} & 17.5GB \\
    \rowcolor{drawioblue}
    + \textbf{ISAC$_\text{LO}$ (Ours)} & \textbf{100\%} & \textbf{90\%} & \textbf{51\%} & \textbf{40\%} & \textbf{70\%} & 21s & {9.7GB} \\

    \midrule
    SDXL \cite{podell2023sdxl}  & 90\% & 71\% & 49\% & 32\% & 61\% & \textbf{48s} & \textbf{12.8GB} \\
    + CountGen {\tiny\color{gray}CVPR'25}\cite{binyamin2024count}  & \textbf{97\%} & {83\%} & {52\%} & {44\%} & {69\%} & {100s} & 55.3GB \\
    \rowcolor{drawioblue}
    + \textbf{ISAC$_\text{LO}$ (Ours)}  & {96\%} & \textbf{89\%} & \textbf{71\%} & \textbf{47\%} & \textbf{76\%} & 101s & {29.8GB} \\
    
    \bottomrule
\end{tabular}
}

\includegraphics[width=\linewidth]{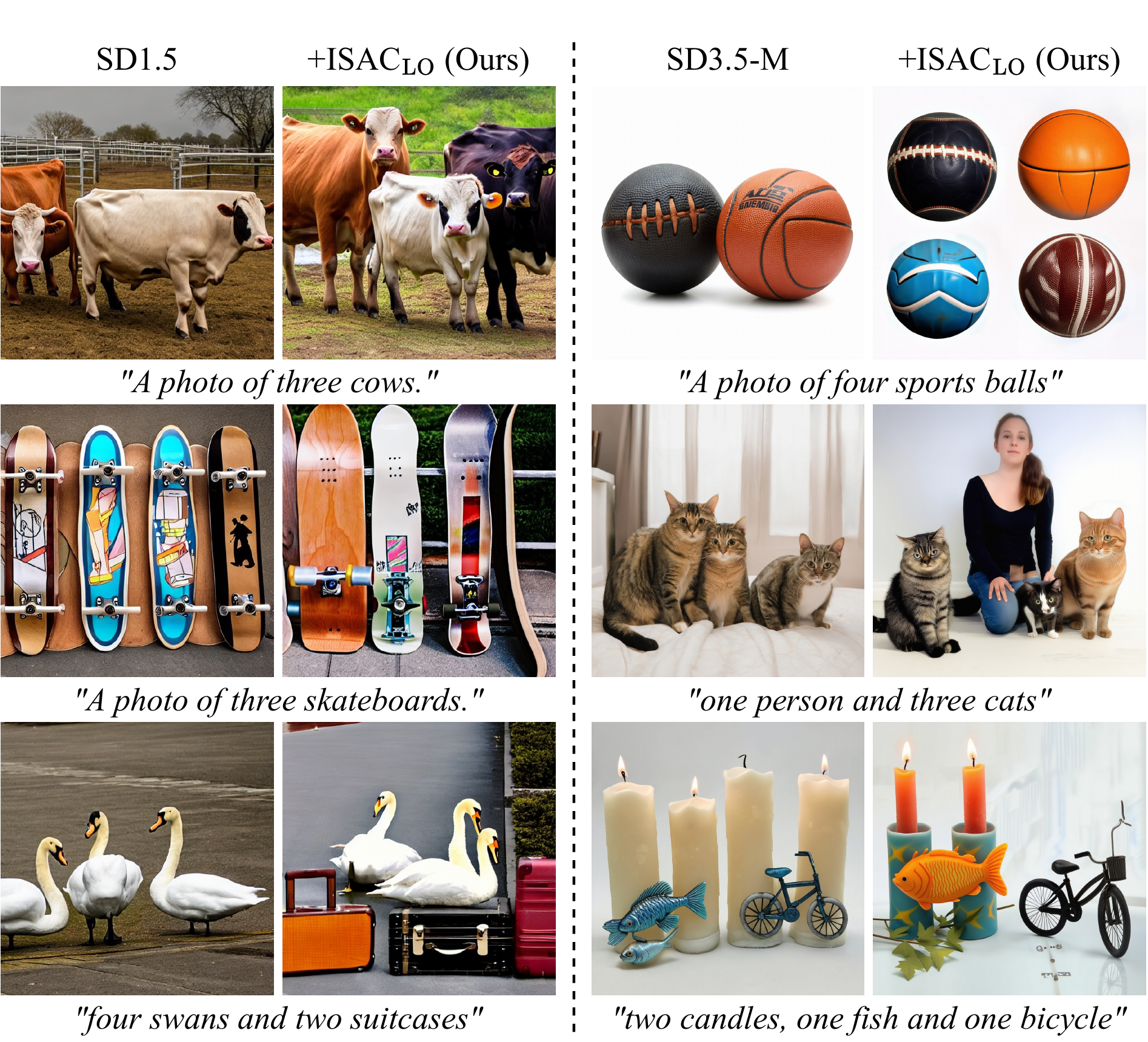}
\captionof{figure}{
    Qualitative results of ISAC$_\text{LO}$ for exact-instance-count prompts on SD1.5 \cite{rombach2022high} and SD3.5-M \cite{esser2024scaling}.
}
\label{fig:qual_count}
\end{minipage}
\hfill %
\begin{minipage}[t]{0.46\linewidth}
\centering

\captionof{table}{Application of ISAC$_\text{LO}$ to layout-to-image generation.}
\label{tab:layout_quantitative}
\vspace{1.2em}

\setlength{\tabcolsep}{3pt}       

\newcolumntype{C}[1]{>{\centering\arraybackslash}p{#1}}
\setlength{\mycolwidth}{0.18\textwidth} 

\resizebox{\linewidth}{!}{
\begin{tabular}{
    l   
    *{4}{C{\mycolwidth}}  
}
    \toprule
    \multirow{2}{*}{Method}
      & \multicolumn{4}{c}{HRSBench (\(\uparrow\))} \\
    \cmidrule(lr){2-5}
      & Counting (F1)
      & Color (Acc.)
      & Spatial (Acc.)
      & Size (Acc.) \\
    \midrule
    
    GLIGEN \cite{li2023gligen}                & 0.666 & 0.307 & 0.268 & 0.188 \\
    + CAR\&SAR \cite{phung2024grounded} &  0.675 & 0.402 & 0.277 & 0.263 \\
    \rowcolor{drawioblue}
    + \textbf{ISAC$_\text{LO}$ (Ours)}                      & \textbf{0.713} & \textbf{0.452} & \textbf{0.281} & \textbf{0.275} \\
    \bottomrule
\end{tabular}%
}

\vspace{0.8em}
\includegraphics[width=\linewidth]{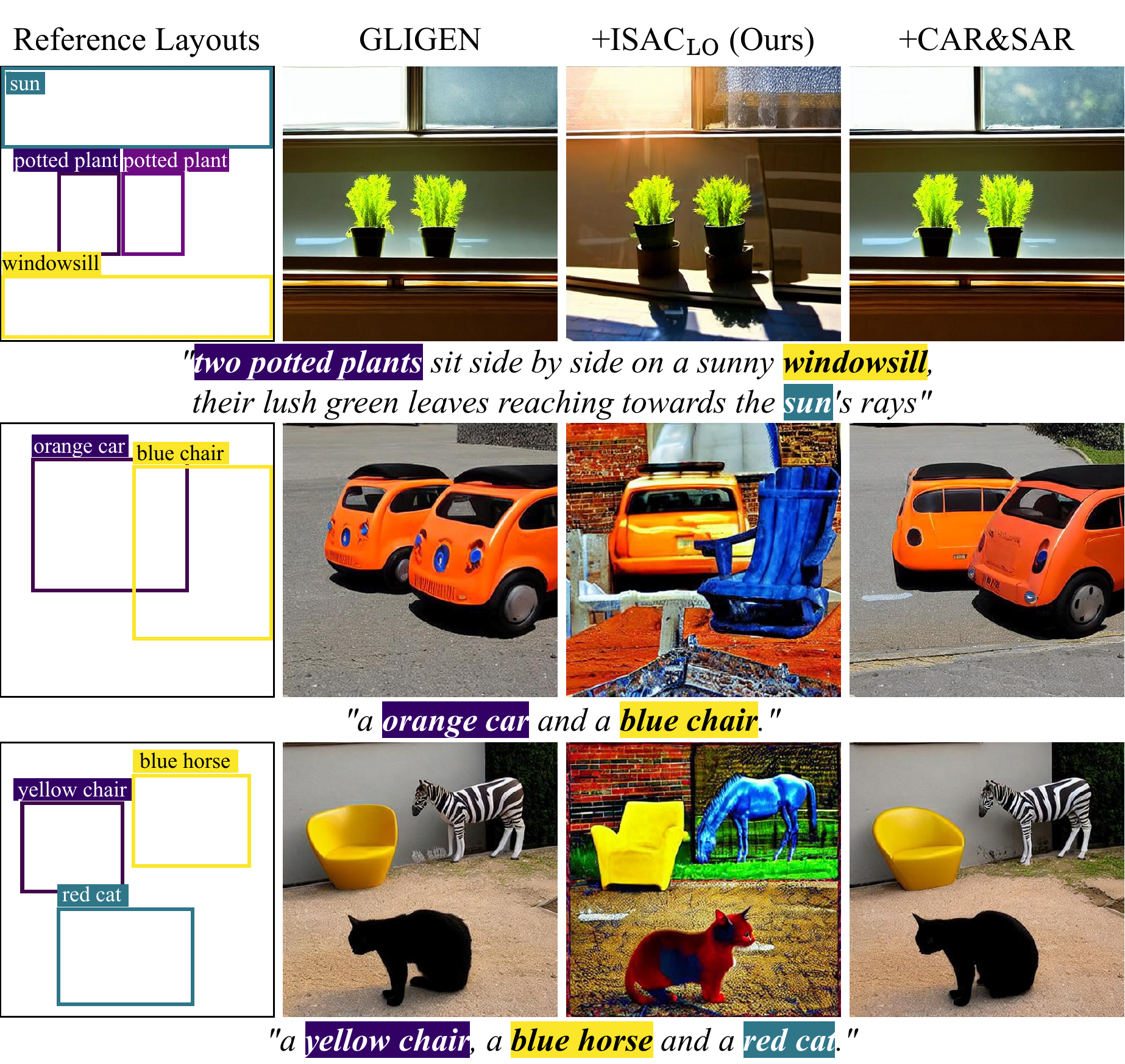}
\captionof{figure}{
Qualitative results of ISAC$_\text{LO}$ with the GLIGEN~\cite{li2023gligen} layout-to-image controller.
}
\label{fig:qual_layout}
\end{minipage}
\end{table*}

\subsubsection{ISAC vs.\ prior count-supervised methods.} 
As shown in \cref{tab:count_results,fig:qual_count}, ISAC$_\text{LO}$ achieves 70\% and 76\% instance-counting accuracy on SD1.4~\cite{rombach2022high} and SDXL~\cite{podell2023sdxl}, surpassing count-supervised methods (49\% for Counting Guidance~\cite{kang2025counting} and 69\% for CountGen~\cite{binyamin2024count}) even though all methods are given the same instance-count supervision and ISAC$_\text{LO}$ uses no fine-tuning or extra training data. Counting-based supervision for auxiliary models can only be exploited once semantic evidence is sufficiently clear, so its effect is limited to later diffusion steps and to samples where object categories are already well formed. By contrast, ISAC$_\text{LO}$ strengthens instance structure at early timesteps, before semantics have fully emerged, by reshaping attention to separate and stabilize instance-level regions; this enables it to reach the highest counting accuracy without auxiliary networks~\cite{hobley2022learning} or extra labels such as instance-level mask annotations.

\subsubsection{Task flexibility of ISAC.}
Beyond the text-to-image setting in \cref{tab:main-results,tab:count_results}, \cref{tab:layout_quantitative} shows that ISAC$_\text{LO}$ also improves layout-to-image controllers \cite{li2023gligen}. On HRS-Bench, ISAC$_\text{LO}$ yields the largest gains compared with layout refinements \cite{phung2024grounded}. This advantage comes from enforcing instance separation for adjacent boxes early in the diffusion trajectory, rather than only constraining attention within each box. 
By carving out dense instance masks from initial coarse box layouts, 
ISAC$_\text{LO}$ prevents neighboring objects from being merged and delivers more reliable counting in crowded layouts as shown in \cref{fig:qual_layout}. \Cref{fig:qual_more_layout} in Appendix~\ref{appendix_qualitative} provides more results on layout tasks.

\subsubsection{Scalability via latent selection.}
\label{subsubsec:latent-selection}
\begin{figure}[t]
    \centering
    \includegraphics[width=.7\linewidth]{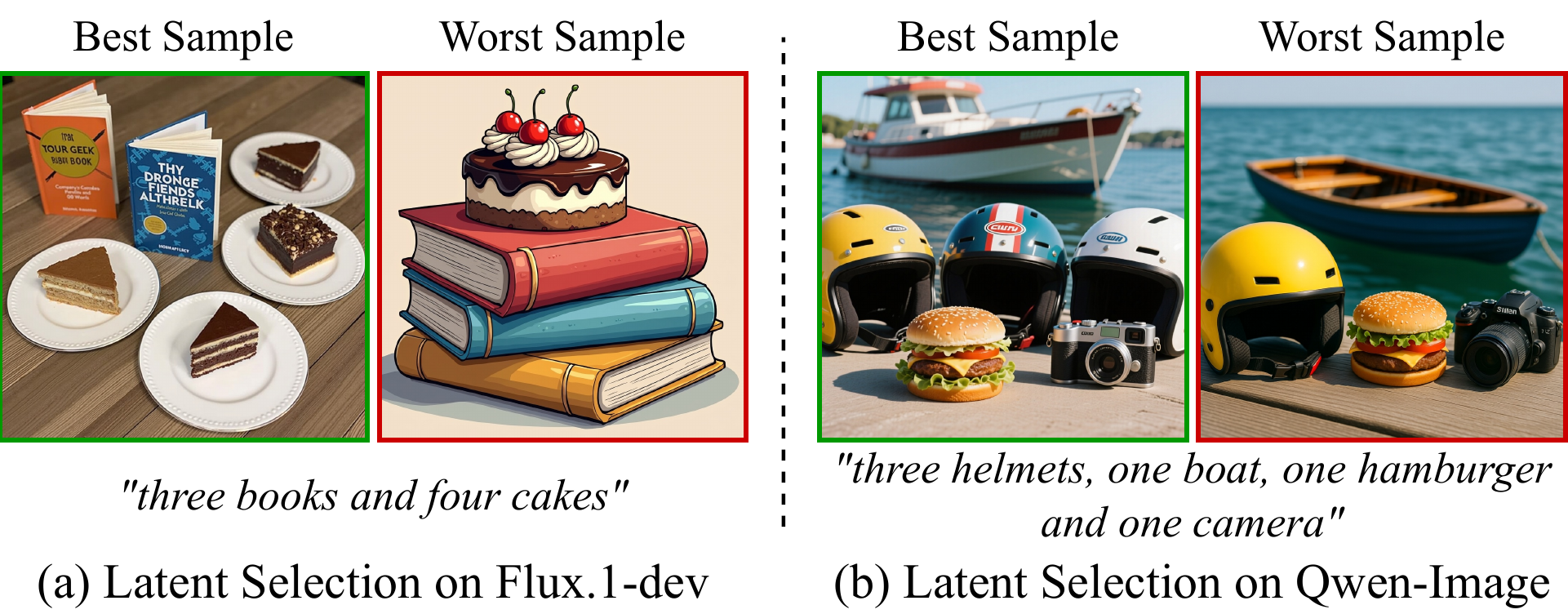}
    \caption{
    Qualitative results of latent selection with ISAC$_\text{LS}$ scoring.
    }
    \label{fig:qual_selection}
\end{figure}

In terms of computational cost, ISAC$_\text{LO}$ lies in the same latency increase as prior latent-optimization attention-control methods, while incurring the expected additional VRAM usage from backpropagation (\cref{tab:main-results}).
For larger backbones, such as Flux \cite{flux2024,flux-2-2025} and Qwen-Image \cite{wu2025qwenimagetechnicalreport}, ISAC$_\text{LS}$ (\cref{alg:isac_selection}) provides a gradient-free alternative that avoids backpropagation and uses the ISAC objective only as a verifier.  
\Cref{tab:isac-selection-result} shows the effectiveness of ISAC$_\text{LS}$ on multi-object generation. \Cref{fig:qual_selection} further shows that the ISAC$_\text{LS}$ score distinguishes samples with missing instances or semantic mixing from better candidates.

\begin{algorithm}[b]
\caption{ISAC with Latent Selection (ISAC$_\text{LS}$)}
\label{alg:isac_selection}
\DontPrintSemicolon

\KwIn{Prompt $\mathcal{T}$, Model $\epsilon_\theta$, decoder $\mathcal{D}$, Batch size $B$}
\KwOut{Image $I_0$}
$X^{(i)}_T \sim \mathcal{N}(0,I)$, $S[i]=0,\;\forall i=1,\dots,B$ \;
\For{$i=1,\dots,B$}{
  \For{$t=T,T{-}1,\ldots,1$}{
  $X^{(i)}_{t-1} \!\gets\! \text{Denoise}(X^{(i)}_t, \mathcal T,\epsilon_\theta, t)$ \\ \qquad$\texttt{with Hooks} \to SA^{(i)}_t, CA^{(i)}_t$\;
    $CA_t^{{(i)},\text{ins}} \gets SA^{(i)}_t \cdot CA^{(i)}_t$ \;
    \texttt{Build} foreground gate and instance masks (Eqs. \ref{eq:binarize}, \ref{eq:fg_mask}, \ref{eq:instance_mask}) \;
    \texttt{Compute }$\mathcal{L}_\text{ins},\mathcal{L}_\text{sem} $ (Eqs. \ref{eq:L_ins}, \ref{eq:L_sem}) \;
    $\mathcal{L}_\text{ISAC}(X^{(i)}_t,t) \gets \lambda_\text{ins}(t)\mathcal{L}_\text{ins}(X^{(i)}_t) + \lambda_\text{sem}(t)\mathcal{L}_\text{sem}(X^{(i)}_t)$\;
    \texttt{Score Update:} $S[i]\gets S[i] + \mathcal{L}_\text{ISAC}(X^{(i)}_t,t)$
  }
}
$i^\star=\arg \min _{i}S[i]$  \tcp*{Best scored latent}
$I_0 \gets \mathcal{D}(X^{(i^\star)}_0)$ \tcp*{Decode to pixel space}
\end{algorithm}

\begin{table*}[t]
\centering
\setlength{\tabcolsep}{3pt}       

\caption{\textbf{Best-of-10 latent selection on IntraCompBench.} The strategy is applied with ISAC using the base models~\cite{flux2024,flux-2-2025,wu2025qwenimagetechnicalreport}.}
\label{tab:isac-selection-result}
\newcolumntype{C}[1]{>{\centering\arraybackslash}p{#1}}

\newlength{\selectioncolwidth}
\setlength{\selectioncolwidth}{0.06\textwidth} 

\resizebox{\linewidth}{!}{
\begin{tabular}{
    l   
    *{5}{C{\selectioncolwidth}}  
    *{5}{C{\selectioncolwidth}}  
    c   
    c   
}
    \toprule
    \multirow{2}{*}{Method}
      & \multicolumn{5}{c}{Multi‑Class Accuracy (\(\uparrow\))}
      & \multicolumn{5}{c}{Multi‑Instance Accuracy (\(\uparrow\))}
      & \multicolumn{2}{c}{Efficiency (\(\downarrow\))} \\
    \cmidrule(lr){2-6} \cmidrule(lr){7-11} \cmidrule(lr){12-13}
      & \#2 & \#3 & \#4 & \#5 & Avg.
      & \#2 & \#3 & \#4 & \#5 & Avg.
      & Latency & VRAM \\
    \midrule

    \(\text{Flux.1-dev}\) \cite{flux2024}     & 84\% & 37\% & 3\% & 2\% & 31\% & 97\% & 89\% & 82\% & 66\% & 83\% & \textbf{50s} & \textbf{37.2GB} \\
    \rowcolor{drawioblue}
    + \(\textbf{ISAC}_\text{LS}\) \textbf{(Ours)} & \textbf{97\%} & \textbf{48\%} & \textbf{38\%} & \textbf{19\%} & \textbf{51\%} & \textbf{99\%} & \textbf{94\%} & \textbf{85\%} & \textbf{72\%} & \textbf{88\%} & 85s & 40.8GB \\
    \midrule

    \(\text{Qwen-Image}\) \cite{wu2025qwenimagetechnicalreport} & 91\% & 45\% & 33\% & 10\% & 48\% & 98\% & 92\% & 84\% & 70\% & 86\% & \textbf{140s} & \textbf{60.1GB} \\
    \rowcolor{drawioblue}
    + \(\textbf{ISAC}_\text{LS}\) \textbf{(Ours)}& \textbf{99\%} & \textbf{58\%} & \textbf{42\%} & \textbf{25\%} & \textbf{56\%} & \textbf{99\%} & \textbf{96\%} & \textbf{89\%} & \textbf{78\%} & \textbf{91\%} & 210s & 65.3GB \\
    
    \midrule
    \(\text{Flux.2-dev}\) \cite{flux-2-2025} & 97\% & 95\% & 84\% & 78\% & 88\% & \textbf{100\%} & 93\% & 81\% & 75\% & 87\% & \textbf{205s} & \textbf{74.2GB} \\
    \rowcolor{drawioblue}
    + \(\textbf{ISAC}_\text{LS}\) \textbf{(Ours)} & \textbf{99\%} & \textbf{98\%} & \textbf{89\%} & \textbf{83\%} & \textbf{92\%} & \textbf{100\%} & \textbf{98\%} & \textbf{88\%} & \textbf{81\%} & \textbf{92\%} & 305s & 79.8GB \\
    
    \bottomrule
    \end{tabular}%
}
\end{table*}

\subsubsection{Applicability to few-step models.}
A key concern is whether ISAC$_\text{LO}$ remains effective under the limited inference budgets of few-step models. We validate our method on Z-Image-Turbo~\cite{team2025zimage} and Flux.2-klein-4B~\cite{flux2klein4b}, which utilize only 8 and 4 steps, respectively. The results in \cref{tab:isac-fewstep-result} demonstrate that ISAC$_\text{LO}$ effectively boosts multi-object synthesis even with fewer opportunities for diffusion guidance. This confirms ISAC's suitability for low-latency applications.

\begin{table*}[t!]
\centering
\setlength{\tabcolsep}{3pt}       

\caption{Application of ISAC$_\text{LO}$ on few-step text-to-image diffusion models.
}
\label{tab:isac-fewstep-result}

\newcolumntype{C}[1]{>{\centering\arraybackslash}p{#1}}

\newlength{\fewstepcolwidth}
\setlength{\fewstepcolwidth}{0.06\textwidth} 

\resizebox{\linewidth}{!}{
\begin{tabular}{
    l   
    *{5}{C{\fewstepcolwidth}}  
    *{5}{C{\fewstepcolwidth}}  
    c   
    c   
}
    \toprule
    \multirow{2}{*}{Method}
      & \multicolumn{5}{c}{Multi‑Class Accuracy (\(\uparrow\))}
      & \multicolumn{5}{c}{Multi‑Instance Accuracy (\(\uparrow\))}
      & \multicolumn{2}{c}{Efficiency (\(\downarrow\))} \\
    \cmidrule(lr){2-6} \cmidrule(lr){7-11} \cmidrule(lr){12-13}
      & \#2 & \#3 & \#4 & \#5 & Avg.
      & \#2 & \#3 & \#4 & \#5 & Avg.
      & Latency & VRAM \\
    \midrule

    \(\text{Z-Image-Turbo}\) \cite{team2025zimage}     & 67\% & 58\% & 38\% & 32\% & 48\% & \textbf{100\%} & 92\% & 45\% & 37\% & 68\% & \textbf{6s} & \textbf{33GB} \\
    \rowcolor{drawioblue}
    + \(\textbf{ISAC$_\text{LO}$}\) \textbf{(Ours)} & \textbf{87\%} & \textbf{74\%} & \textbf{61\%} & \textbf{49\%} & \textbf{68\%} & \textbf{100\%} & \textbf{97\%} & \textbf{62\%} & \textbf{49\%} & \textbf{77\%} & 13s & 78GB \\
    \midrule

    \(\text{Flux.2-klein-4B}\) \cite{flux2klein4b}     & 79\% & 45\% & 12\% & 1\% & 34\% & 98\% & 81\% & 75\% & 49\% & 75\% & \textbf{4s} & \textbf{29GB} \\
    \rowcolor{drawioblue}
    + \(\textbf{ISAC$_\text{LO}$}\) \textbf{(Ours)} & \textbf{89\%} & \textbf{63\%} & \textbf{43\%} & \textbf{19\%} & \textbf{54\%} & \textbf{100\%} & \textbf{93\%} & \textbf{86\%} & \textbf{65\%} & \textbf{86\%} & 8s & 64GB \\

    \bottomrule
    \end{tabular}%
}
\end{table*}

\subsubsection{Importance of instance-to-semantic schedule.} Table \ref{tab:ablation-schedule} shows that both losses and their ordering are critical. Optimizing only the instance term (A) yields strong multi-instance accuracy but almost no gain on multi-class prompts, indicating that it can form the right number of structures but cannot reliably assign semantics. Using only the semantic term or a fixed balance (B–C) is also suboptimal, since semantic separation without prior boundary stabilization is unstable. The reverse schedule that goes from semantic to instance (D) further degrades multi-class accuracy. Our instance-to-semantic schedule (E) achieves the best performance on both metrics, supporting the hypothesis that instance structure should be established first and then refined semantically.

\begin{table}[t]
\centering
\setlength{\tabcolsep}{6pt}       
\caption{\textbf{Effect of the loss schedule on ISAC$_\text{LO}$.} ``Class'' and ``Instance'' denote multi-class and multi-instance accuracy on IntraCompBench, respectively.}
\label{tab:ablation-schedule}
\resizebox{0.65\linewidth}{!}{
\begin{tabular}{l l c c c c}
    \toprule
            & Description
            & $\lambda_{\text{ins}}(t)$
            & $\lambda_{\text{sem}}(t)$
            & Class 
            & Instance \\ 
    \midrule
    A & Instance Only        & 1         & 0         & 10\% & 65\% \\
    B & Semantic Only        & 0         & 1         & 28\% & 54\% \\
    C & Fixed Balance        & 0.5       & 0.5       & 25\% & 60\% \\
    D & Semantic-to-Instance & $1 - t/T$ & $t/T$     & 21\% & 55\% \\
    \rowcolor{drawioblue}
    E & Instance-to-Semantic & $t/T$     & $1 - t/T$ & \textbf{36\%} & \textbf{69\%} \\
    \bottomrule
\end{tabular}%
}
\end{table}

\section{Conclusion}

In this work, we study multi-instance generation in diffusion models through the interaction between instance structure and token-conditioned semantics. We show that count failures and semantic mixing arise when semantic binding proceeds before instance boundaries are reliably stabilized. Building on this diagnosis, we introduce ISAC, a training-free, model-agnostic objective that first stabilizes instance regions from structural cues and then binds classes and attributes within each region. 
Across three benchmarks and multiple diffusion backbones, ISAC improves both text-to-image and layout-to-image generation. On IntraCompBench, it achieves higher counting and compositional accuracy than prior count-supervised approaches without additional data, fine-tuning, or external vision models. 
Overall, instance-first control provides a practical, reproducible path toward narrowing the multi-instance reliability gap of open-weight diffusion models. It further suggests a promising direction for decoupled structure--semantics control in video and other structured generative settings.

\section*{Acknowledgments}
This work was partly supported by the KHIDI grant funded by the Korean government (MOHW) [No.RS-2025-02307233], the NRF or IITP grants funded by the Korean government (MSIT) [RS-2026-25507282 (5\%), RS-2026-25518317 (5\%), No.RS-2024-00405857(10\%), No.05-26-04-0094, No.RS-2026-25472075, No.RS-2026-25483206, No.RS-2025-02305581, No.RS-2025-25442338, and No.RS-2021-II211343(5\%)], the ITIP grant funded by the Korean government (MOTIR) [No.RS-2026-25549946], the Research grant from SNU, and the Strategic Hub grant for International Research Collaboration of SNU.

Kyungsu Kim is affiliated with the School of Transdisciplinary Innovations, Department of Biomedical Science, Interdisciplinary Program in Artificial Intelligence (IPAI), Medical Research Center, and AI Institute at SNU.

%
%
\bibliographystyle{splncs04}
\bibliography{main}

\clearpage
\appendix
\renewcommand{\theHsection}{appendix.\arabic{section}}

\section{Implementation Details}
\label{sec:implementation_details}

\subsection{Method Details}

\subsubsection{Attention accumulation.}
We accumulate self- and cross-attention maps from all attention layers. When self-attention (SA) and cross-attention (CA) maps have different spatial resolutions across layers, in case of U-Net~\cite{ronneberger2015u} architectures, we bilinearly upsample each map to the highest spatial resolution \((H\times W)\) and average over layers and heads to obtain a single SA/CA pair \emph{per step}:
\begin{align}
\label{eq:attn-accum}
    &SA_t = \frac{1}{\mathcal N} \sum_{l=1}^{M} \sum_{h=1}^{h_l}\texttt{Upsample}(SA_l^h(X_t), \delta_l)\\ 
    &CA_t = \frac{1}{\mathcal N} \sum_{l=1}^{M} \sum_{h=1}^{h_l}\texttt{Upsample}(CA_l^h(X_t,\mathcal{T}), \delta_l)\!
\end{align}
where \(\delta_l = H/H_l = W/W_l\) is the upsampling factor for layer \(l\), and \(\mathcal N = \sum_{l=1}^{M} h_l\) is the total number of heads.
Then, we apply $\texttt{min-max}$ normalization to the accumulated maps as defined in \cref{eq:minmax_norm}. For a fair comparison, we use the same attention accumulation scheme across all baseline methods \cite{rombach2022high, podell2023sdxl, chen2023pixartalpha, chen2024pixart, esser2024scaling, flux2024, wu2025qwenimagetechnicalreport}.

\subsubsection{Choice of the step size $\eta$.}
We evaluate the sensitivity of ISAC$_\text{LO}$ to the gradient step size $\eta$ in \cref{alg:isac_optimization} under our instance-to-semantic schedule. When $\eta$ is too small, the optimization loss barely decreases and has negligible impact on the generated images. In contrast, excessively large values of $\eta$ cause latent collapse and noticeably degrade image quality. This trade-off is consistent with other latent-optimization methods \cite{chefer2023attendandexcite}, and ISAC$_\text{LO}$ is subject to the same limitation.

\begin{figure}[b]
    \centering
    \includegraphics[width=.6\linewidth]{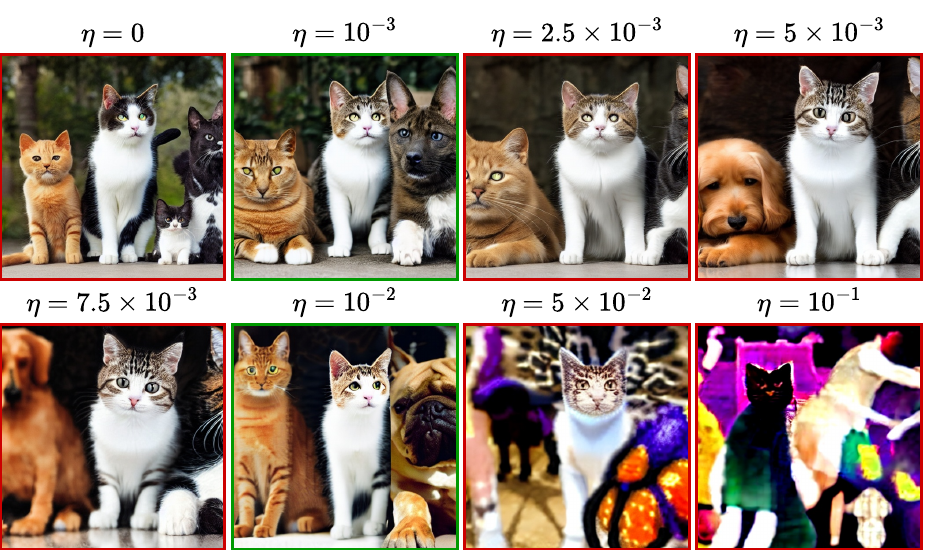}
    \caption{Qualitative comparison of ISAC$_\text{LO}$ with various step sizes $\eta$. For all cases, we provide \textit{``A photo of two cats and a dog''} as the input prompt and use SD1.5 \cite{rombach2022high} as the baseline diffusion model.}
    \label{fig:LR_sensitivity}
\end{figure}

Empirically, we find that a step size of $\eta = 10^{-2}$ achieves stable improvement without visible artifacts across backbones and datasets (\cref{fig:LR_sensitivity}). The gradient-descent update $\tilde{X}_t \gets X_t - \eta \nabla_{X_t}\mathcal{L}_t(X_t)$ in \cref{alg:isac_optimization} is applied once per diffusion timestep, and $\eta$ is the only additional hyperparameter introduced by ISAC$_\text{LO}$; all other settings follow the default configuration of each backbone.

\subsubsection{Foreground-gate.} 
\begin{figure}[t]
    \centering
    \includegraphics[width=0.8\linewidth]{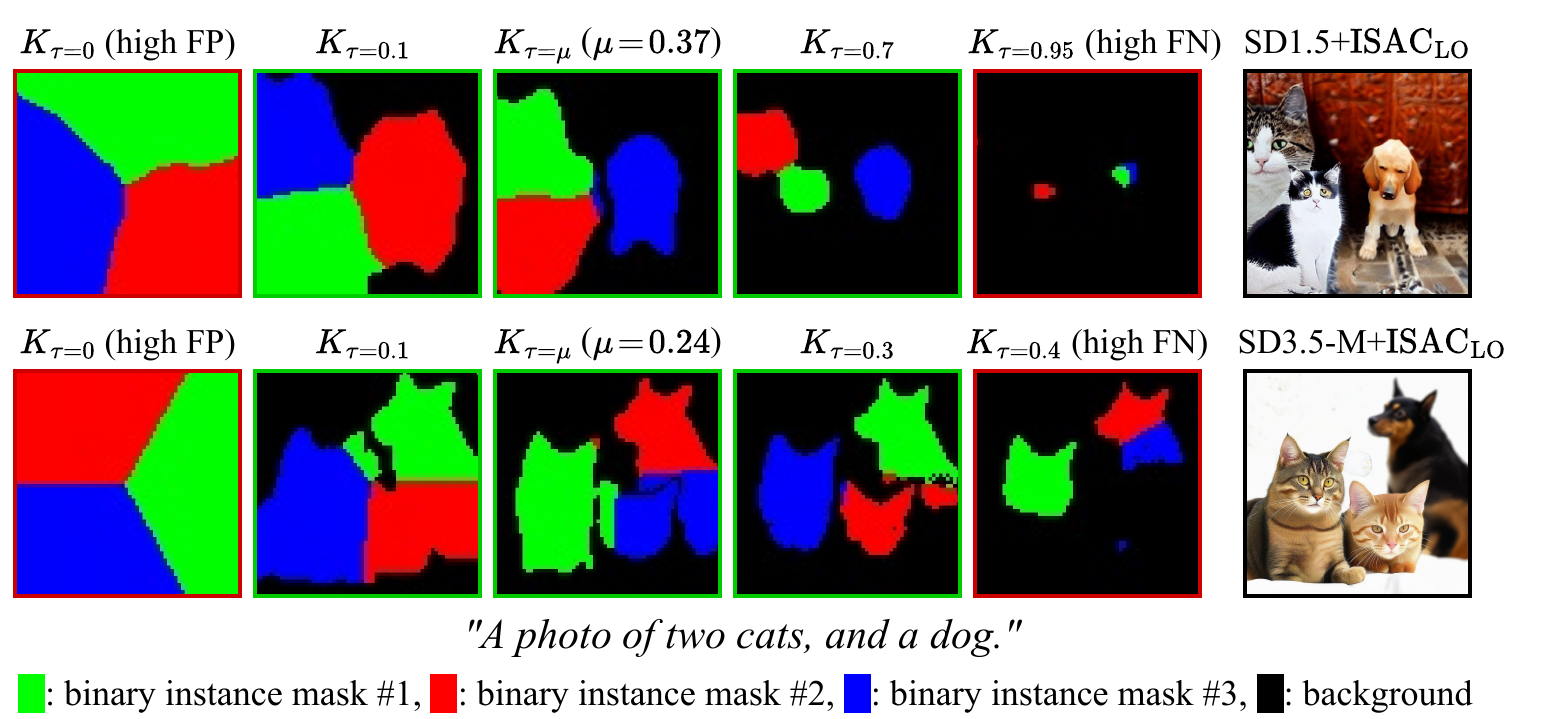}
    \caption{
    \textbf{Foreground-gate threshold analysis.} 
    Binary instance masks \(K\) are extracted at \(t=T/2\) as in \cref{fig:method-overview} under different thresholds \(\tau\) on SD1.5 (top) and SD3.5-M (bottom).
    The rightmost images are generated by \(\text{ISAC}_\text{LO}\) with mean thresholding \((\tau=\mu)\).
    }
    \label{fig:fg_gate_failure_analysis}
\end{figure}

We restrict self-attention clustering to foreground regions to reduce potential clustering errors caused by background pixels. We use an adaptive foreground threshold, defined as the mean value of $CA_t^\text{ins}$ (\cref{eq:binarize}), rather than a fixed threshold.
\Cref{fig:fg_gate_failure_analysis} supports the design choice. Low threshold $\tau$ admits background (FP), while high $\tau$ drops foreground (FN). Our adaptive mean threshold ($\tau=\mu$) scales with per-timestep attention magnitude, avoiding both. Still, small or thin objects remain challenging.

\subsubsection{Details of clustering with coordinates.} 
\begin{figure}[t]
  \centering
  \includegraphics[width=0.6\linewidth]{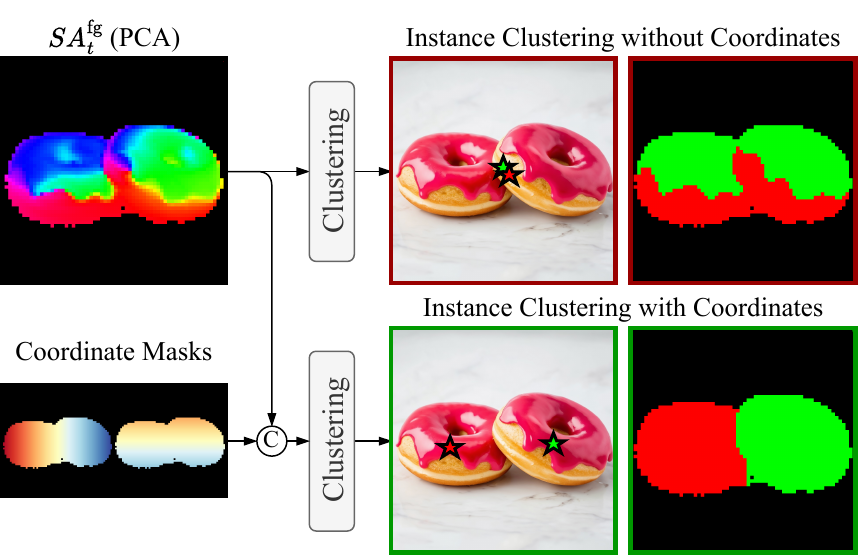}
  \caption{
  Concatenating normalized coordinates $(x,y) \in [-1,1]^2$ to SA features stabilizes clustering, reducing erroneous merges and making subtle boundary cues more salient.
  The image is generated with the prompt, \textit{``a photo of two donuts''} on SD3.5-M~\cite{esser2024scaling}. 
  We averaged all accumulated self/cross-attention maps over timesteps into a single (SA, CA) pair. Then, clustering algorithms are applied to those maps with or without concatenation of spatial coordinates.
  Note that this is commonly used in previous literature~\cite{comaniciu2002mean,hwang2019segsort,liu2018intriguing}.
  }
  \label{fig:coord-concat}
\end{figure}

We visualize how and why normalized coordinates $(x,y) \in [-1,1]^2$ are concatenated to the feature vectors in \cref{eq:instance_mask}. When adjacent instances are highly semantically similar, their foreground positions can have similar SA features, and weak boundary cues are easily overwhelmed (\cref{fig:coord-concat}, top/red). By forming a joint space of SA features and spatial coordinates and performing clustering in this augmented space, the clustering better respects spatial connectivity within each instance, making boundaries more separable and making subtle boundary cues more salient (\cref{fig:coord-concat}, bottom/green). 

This spatial augmentation is also quantitatively important. Clustering with SA features alone yields 21\% multi-class accuracy, whereas the default SA+coordinate clustering achieves 36\%, indicating that coordinates act as a spatial regularizer for separating adjacent objects.

\subsubsection{Choice of clustering algorithm.} 
\begin{figure}[t]
    \centering
    \includegraphics[width=.6\linewidth]{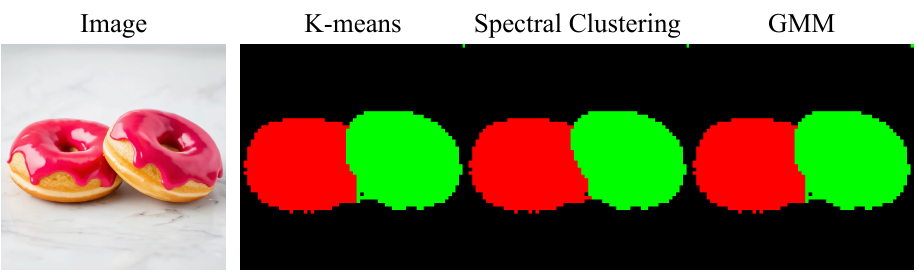}
    \caption{Qualitative comparison of clustering algorithms. 
    The image is generated with the prompt, \textit{``a photo of two donuts''}. 
    We averaged all accumulated self/cross-attention maps over timesteps into a single (SA, CA) pair. Then, clustering algorithms are applied to those maps with concatenation of spatial coordinates. This is an extension of \cref{fig:coord-concat}.
    }
    \label{fig:clustering}
\end{figure}

We compare alternative clustering algorithms in \cref{fig:clustering} and \cref{tab:clustering_alternatives}. While K-means, spectral clustering, and Gaussian mixture models (GMMs) achieve similar performance on both \emph{Multi-Class} and \emph{Multi-Instance} accuracy, K-means, which we adopt as our default choice, is approximately $3\times$ to $16\times$ faster. Given this favorable speed–accuracy trade-off, we use K-means for ISAC.

\begin{table}[t!]
\centering
\setlength{\tabcolsep}{4pt}       

\caption{Comparison of clustering algorithms in terms of accuracy (\%) and latency (ms). Latency is defined as the execution time for a one-time application of each clustering algorithm to a self-attention map. This table can be seen as an extension of \cref{tab:ablation-schedule}.}
\label{tab:clustering_alternatives}

\resizebox{0.8\linewidth}{!}{
\begin{tabular}{lccc}
    \toprule
    Clustering Algorithm 
        & Multi-Class (\(\uparrow\))
        & Multi-Instance (\(\uparrow\))
        & Latency (\(\downarrow\))\\
    \midrule
    \rowcolor{drawioblue}
    \textbf{K-means}    & \textbf{36\%}                           & \textbf{69\%}                              & \textbf{505 ms}                \\
    Spectral Clustering           & 35\%                           & \textbf{69\%}                              & 1,630 ms               \\
    GMM       & \textbf{36\%}                           & \textbf{69\%}                              & 8,313 ms                \\
    \bottomrule
\end{tabular}
}
\end{table}


\begin{figure*}[t]
    \centering
    \includegraphics[width=\textwidth]{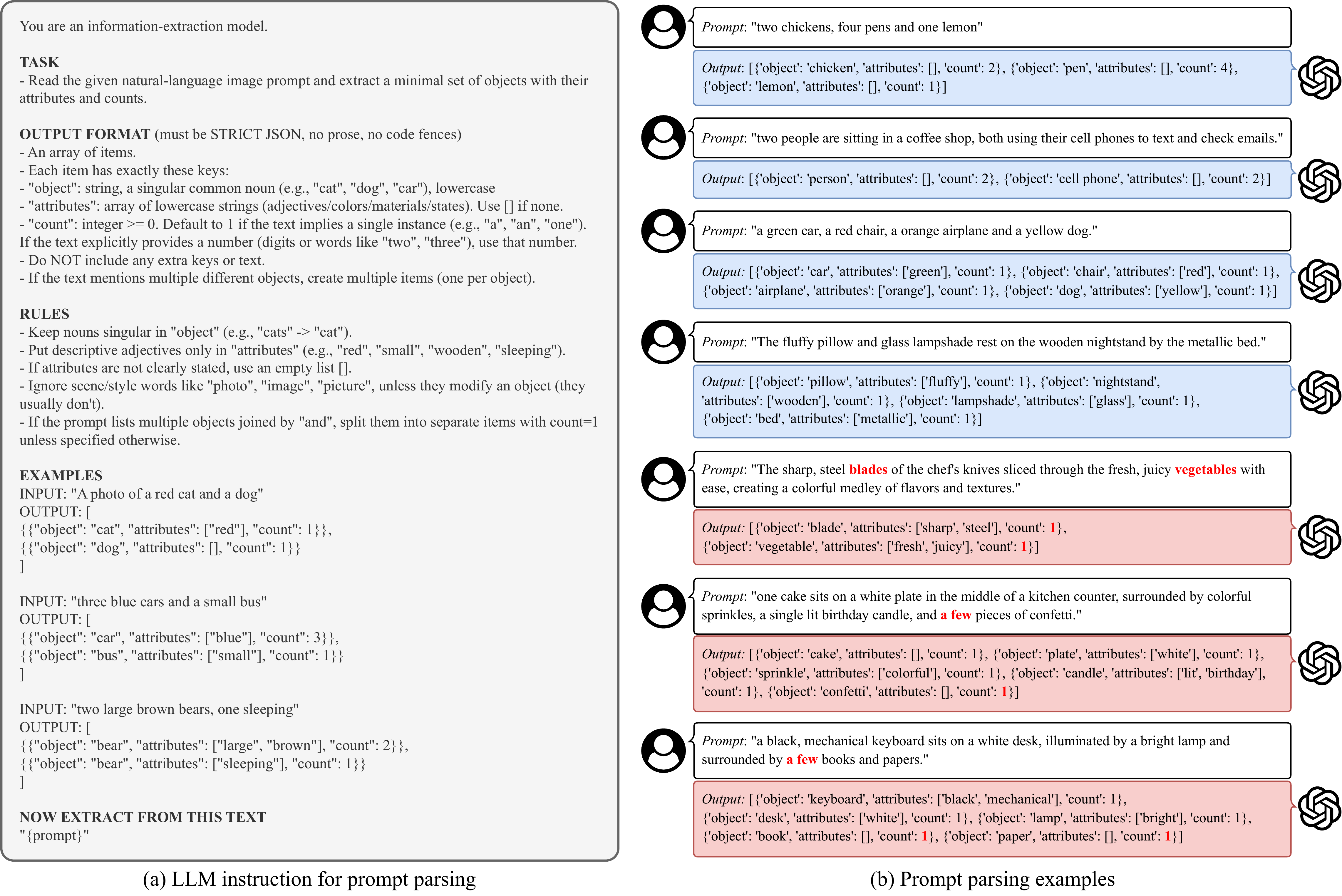}
    \caption{Details on LLM-guided automatic prompt parsing.}
    \label{fig:llm-parser}
\end{figure*}

\subsubsection{Automatic prompt parsing.} We use an LLM-based parser to automatically extract class tokens $\tau_i$, counts $n_i$, attributes $\{\chi_{i,j}\}$, and token relations for $P_{\text{repel}}$ and $P_{\text{bind}}$ from natural-language prompts. For $P_{\text{repel}}$ and $P_{\text{bind}}$, the parser resolves only object-local attribute pairs. If an attribute is global or cannot be uniquely assigned to a noun/class, we conservatively omit the bind pair. \cref{fig:llm-parser}(a) shows the instruction used for this parsing procedure and \cref{fig:llm-parser}(b) provides example outputs on prompts from T2I-CompBench and HRS-Bench. In all experiments, we use GPT-OSS-20B \cite{openai2025gptoss120bgptoss20bmodel}, which produces reliable and consistent parses in practice. For a moderately complex prompt such as \textit{``The fluffy pillow and glass lampshade rest on the wooden nightstand by the metallic bed.''}, a single forward pass of the parser takes about 10 seconds and requires roughly 40 GB of VRAM on a single A100 GPU. 

While the LLM parser produces highly accurate outputs overall, our current rules do not resolve ambiguous count expressions such as ``a few'', ``several'', ``a couple of'', or plurals without explicit numerals (\eg, ``blades''). Handling these ambiguous counts is beyond the scope of our experiments, but it could be addressed by sampling counts from a small candidate set, for example $\mathcal{N} = \{3,4\}$, or by providing targeted few-shot examples to the LLM.

\subsubsection{Cross-attention normalization.}
Baseline methods, Attend-and-Excite \cite{chefer2023attendandexcite}, InitNO \cite{guo2024initno}, and TEBOpt \cite{Chen_2024_TEBOpt}, use the \texttt{softmax} normalization technique on cross-attention maps. It operates by applying the \texttt{softmax} function to the cross-attention maps along the token dimension, excluding the \texttt{SOT} token at index 0. This sharpens the attention, emphasizing foreground objects while suppressing background noise. The formulation is given by:
\begin{equation}
{CA}^\text{softmax}_t = \texttt{softmax}(\beta \cdot CA_t[1:] ).
\end{equation}

Following the official implementation, we set \(\beta=100\) for SD1.4, SD1.5, SD2.1 \cite{rombach2022high} and SD3.5-M \cite{esser2024scaling} across baseline methods through our experiments (see \cref{subsec:main_results,sec:add-quantitative-results}).
In contrast, ISAC adopts a simple element-wise $\texttt{min-max}$ normalization to rescale attention maps:
\begin{equation}
    {CA}^\text{minmax}_t = \frac{CA_t - \min(CA_t)}{\max(CA_t) - \min(CA_t)}.
\label{eq:minmax_norm}
\end{equation}
Although both $\texttt{softmax}$ and $\texttt{min-max}$ normalization are common, we opt for the $\texttt{min-max}$ approach due to its practicality as a parameter-free method.

\subsection{IntraCompBench Details}

Standard benchmarks like T2I-CompBench \cite{huang2025t2ipp}, HRS-Bench \cite{bakr2023hrs} and MultiGen benchmark \cite{Wang2024TokenCompose} do not isolate the \emph{intra‑category} setting where failure on instance discrimination becomes most severe. IntraCompBench is designed to stress this regime and to separately probe the two symptoms we target: (1) \emph{count failures} and (2) \emph{semantic mixing}.

\subsubsection{Task 1 -- multi‑instance accuracy (\%).}
This task isolates count failures. We sample a single class $A$ from a super‑category in \cref{tab:countable-classes} and specify an integer $n\in\{2,3,4,5\}$. Now the prompt is formatted as ``A photo of [n] [class A]s'' (\eg, \textit{``A photo of five cats''}). Success requires producing exactly $N=n$ instances of $A$. This primarily evaluates instance formation.

\subsubsection{Task 2 -- multi‑class accuracy (\%).}
This task stresses semantic mixing. We sample $k\!\in\!\{2,3,4,5\}$ distinct classes within the same super‑category and format into ``A photo of a [class A], a [class B], ..., and a [class E]'' (\eg, \textit{``A photo of a dog, a cat, a horse, a cow, and a sheep''}). Success requires (i) forming $N{=}k$ instances and (ii) assigning the correct semantics to each layout (preventing cross‑object leakage). The intra‑category constraint makes this substantially harder than prior multi‑class settings (\eg, MultiGen \cite{Wang2024TokenCompose}), where inter‑category separability reduces confusion (see \cref{fig:inter_intra_analysis}).

\subsubsection{Class distribution.} 
For reliable automatic evaluation we use a subset of countable object classes from the 80 COCO categories\footnote{Most detection models are trained on COCO classes, so we focus our evaluation on COCO, where the pretrained models are most reliable, rather than experimenting with new classes from datasets like ADE. In practice, MultiGen results in TokenCompose \cite{Wang2024TokenCompose} have shown minimal differences in performance trends between COCO and ADE.} 
\cite{lin2014microsoft}, grouped into four super‑categories: animals, vehicles, sports, and food (\cref{tab:countable-classes}).
We exclude classes such as \textit{``person''}, very small objects (\eg, \textit{``fork''}), ambiguous items (\eg, \textit{``book''}), or objects whose duplication is ill‑defined (\eg, \textit{``bench''}), due to difficulties in reliable detection or instance differentiation.

\begin{table*}[t]
\centering
\setlength{\tabcolsep}{3pt}       
\renewcommand{\arraystretch}{1.2}  

\caption{Countable classes from COCO \cite{lin2014microsoft} dataset used in the evaluation.}
\label{tab:countable-classes}
\begin{tabular}{
    l   
    p{0.85\textwidth}   
}
    \toprule
    Category
      & Classes \\
    
    \midrule
    Animal
      & (9 classes) cat, dog, horse, sheep, cow, elephant, bear, zebra, giraffe \\
    Vehicle
      & (8 classes) bicycle, car, motorcycle, airplane, bus, train, truck, boat \\
    Sports
      & (10 classes) skateboard, snowboard, skis, sports ball, baseball bat, baseball glove, tennis racket, surfboard, kite, frisbee \\
    Food
      & (10 classes) banana, apple, sandwich, orange, broccoli, carrot, hot dog, pizza, donut, cake \\
    \bottomrule
\end{tabular}
\end{table*}

\begin{table}[t]
\centering

\caption{Possible combinations of classes for multi-class evaluation.}
\label{tab:countable-combinations}
\begin{tabular}{
l        
*{4}{c}  
}
    \toprule
    Category
      & \#2 & \#3 & \#4 & \#5 \\
    
    \midrule
    Animal (9 classes)
      & 36 & 84 & 126 & 126 \\
    Vehicle (8 classes)
      & 28 & 56 & 70 & 56 \\
    Sports (10 classes)
      & 45 & 120 & 210 & 252 \\
    Food (10 classes)
      & 45 & 120 & 210 & 252 \\
    \midrule
    Total
      & 154 & 380 & 616 & 686 \\
    \bottomrule
\end{tabular}
\end{table}

\subsubsection{Evaluation via ensemble.} To evaluate accuracy, we use an ensemble of three detectors: Grounding DINO \cite{liu2024grounding}, YOLOE \cite{wang2025yoloerealtimeseeing}, and YOLOv12 \cite{tian2025yolov12}. We adopt a 2-of-3 agreement rule, in other words, an instance is considered correctly detected only if captured by at least two detectors. 
Accuracy is the ratio of correctly detected instances to the target number of instances ($k$ for multi-class, $n$ for multi-instance).
For example, if ``cat'' and ``dog'' are detected in the image from the text prompt ``A photo of a cat, a dog, a horse, a cow and a sheep'', then the accuracy is calculated as \(2 /5=40\%\) for multi-class accuracy. For multi-instance accuracy, if 3 instances of ``cat'' are detected in the image from the text prompt ``A photo of five cats'', then the accuracy is calculated as \(3 /5=60\%\).

\subsubsection{Prompt sampling.}
For multi-class evaluation, we sample 20\% of all combinations in \cref{tab:countable-combinations} for each $k\!\in\!\{2,3,4,5\}$ and generate 10 images per prompt.
For example, in \#5-class evaluation, we randomly sample 25 animal, 11 vehicle, 50 sports, 50 food combinations; 136 prompts total.
For multi-instance evaluation, we generate 10 images for each class in the four super‑categories for every $n\!\in\!\{2,3,4,5\}$, yielding 37 unique prompts per $n$.

\subsection{Evaluation Details}

For the layout-to-image evaluation in \cref{tab:layout_quantitative,fig:qual_layout}, we use HRS-Bench~\cite{bakr2023hrs} and follow the two-stage pipeline established by prior work~\cite{phung2024grounded}: (i) generate text-to-box layouts with an LLM, then (ii) synthesize images conditioned on those layouts. To ensure comparability, we directly reuse the LLM-produced box layouts released in Attention-Refocusing \cite{phung2024grounded} and run only the layout-conditioned image generation step.

Although OverLayBench~\cite{li2025overlaybench} is tailored for evaluating overlapping layouts, its publicly available annotations are not directly compatible with training-free layout-to-image guidance methods~\cite{Xie_2023_ICCV,chen2023trainingfree,xiao2023rb,lee2024groundit,phung2024grounded,dahary2024yourself}. OverLayBench provides a global prompt together with (local prompt, bounding box) pairs, where each local prompt specifies the semantics within its box. Training-based controllers~\cite{li2023gligen,zhou2024migc,wang2024instancediffusion,cheng2024hicohierarchicalcontrollablediffusion,zhou20243dis,zhang2025creatilayout} can ingest an \emph{arbitrary} local prompt via dedicated layout adapters. In contrast, existing training-free layout-to-image guidance methods~\cite{Xie_2023_ICCV,chen2023trainingfree,xiao2023rb,lee2024groundit,phung2024grounded,dahary2024yourself} require per-layout class tokens \(\{\tau_i\}\) to be present in the \emph{global} prompt. Because this condition is not guaranteed in the public OverLayBench data, it precludes a fair comparison with ISAC. We leave a comprehensive quantitative study on OverLayBench---after reconciling prompt formats---to future work.

\clearpage
\section{Broader Related Work Comparisons}

S‑CFG~\cite{shen2024rethinking_SCFG} highlights spatial inconsistencies in global Classifier‑Free Guidance (CFG)~\cite{ho2022classifier} and proposes region‑level guidance that leverages self‑ and cross‑attention maps. However, S‑CFG primarily \emph{amplifies} semantic evidence within semantically segmented patches. It neither disentangles competing semantic signals nor establishes instance boundaries from structural cues.

Contemporary training-free methods also address compositional failures from different perspectives. 
CO3~\cite{dutta2026steer} steers sampling away from mode collisions and improves multi-class composition, but it does not explicitly build spatial instance partitions. 
DeLeaker~\cite{ventura2026deleaker} dynamically reweights attention to mitigate semantic leakage, but, like other token-conditioned
SA-guided methods~\cite{guo2024initno,qiu2025self}, its structural cues remain tied to CA token semantics. As a result, these methods can reduce semantic leakage across different tokens, but they are not designed to separate multiple instances that share the same semantic token. In contrast, ISAC uses CA only to obtain a class-agnostic foreground gate and derives instance partitions directly from SA affinities, decoupling instance discovery from token semantics. These differences among CO3~\cite{dutta2026steer}, DeLeaker~\cite{ventura2026deleaker}, and ISAC are reflected in the IntraCompBench performance results presented in \cref{tab:contemporary_baselines}.

\begin{table}[h]
\centering
\caption{
\textbf{Comparison with contemporary training-free baselines on IntraCompBench.} Multi-Class and Multi-Instance denote average accuracy over the \#2--\#5 settings.}
\setlength{\tabcolsep}{3pt} 

\newcolumntype{C}[1]{>{\centering\arraybackslash}p{#1}}
\newlength{\contempbaselinecolwidth}
\setlength{\contempbaselinecolwidth}{0.12\textwidth} 

\begin{minipage}{0.65\linewidth}
\resizebox{\linewidth}{!}{%
\begin{tabular}{l *{4}{C{\contempbaselinecolwidth}}}
\toprule
  \multirow{2}{*}{Method} & 
  \multicolumn{2}{c}{IntraCompBench ($\uparrow$)} &
  \multicolumn{2}{c}{Efficiency ($\downarrow$)} \\
  \cmidrule(lr){2-3} \cmidrule(lr){4-5}
  & Class & Instance & Latency & VRAM \\
\midrule
SD1.5 \cite{rombach2022high}     & 8\% & 54\% & 8 s & 4.9 GB  \\
$+\ \text{CO3}^\dag$~\cite{dutta2026steer}  & 18\% & 55\% & 23 s & 6.0 GB \\
\rowcolor{drawioblue}
\(+\ \text{ISAC}_\text{LO}\)  & \textbf{36\%} & \textbf{69\%} & 21 s & 9.7 GB  \\
\midrule
SD3.5-M \cite{esser2024scaling}   & 25\% & 64\% & 40 s & 22.9 GB  \\
$+\ \text{DeLeaker}^\ddag$~\cite{ventura2026deleaker}  & 30\% & 62\% & 60 s & 32.0 GB \\
\rowcolor{drawioblue}
\(+\ \text{ISAC}_\text{LO}\)  & \textbf{52\%} & \textbf{83\%} & 140 s & 74.8 GB  \\
\bottomrule
\end{tabular}%
}
\begin{tablenotes}[flushleft]
  \fontsize{7pt}{8pt}\selectfont
  \item $\dag$DDIM-only; not directly applicable to flow-matching backbones.
  \item $\ddag$DiT-only; not directly applicable to U-Net backbones.
\end{tablenotes}
\end{minipage}

\label{tab:contemporary_baselines}
\end{table}

Beyond Counting Guidance~\cite{kang2025counting}, several works~\cite{wang2025training,bansal2023universal} steer generation by querying pretrained vision models as external perceptual signals. Dense Geometry Alignment~\cite{wang2025training} aligns segmentation masks predicted from intermediate images with the target object layouts. Yet, because these vision models~\cite{liu2024grounding} are trained on clean, semantically rich images, their predictions on noisy diffusion states are unreliable, yielding weak guidance (see \cref{fig:dynamics_external_model}). Moreover, many approaches apply predictions on Tweedie‑denoised intermediates~\cite{kim2021noise2score}; Tweedie’s correction is not applicable to flow‑matching models~\cite{lipman2022flow} such as SD3.5‑M~\cite{esser2024scaling}, limiting the practicality of using external models in this regime.

Specifically,
CountGen \cite{binyamin2024count} is a two‑stage approach that first establishes instance‑mask layouts with a fine‑tuned \texttt{ReLayout} module and then applies guidance conditioned on the resulting masks. We find its gains stem largely from the quality of the mask proposals; without explicit instance‑separation guidance, it lags behind ISAC$_\text{LO}$ on our intra‑category settings. ISAC$_\text{LO}$ also complements CountGen: replacing CountGen’s guidance with ISAC$_\text{LO}$ further improves accuracy, indicating that ISAC$_\text{LO}$’s instance discrimination and CountGen’s mask proposals address different parts of the problem (see \cref{tab:count_results_extension}).

\begin{table}[ht]
\centering
\setlength{\tabcolsep}{4pt}       
\caption{
Extended results for CountGen~\cite{binyamin2024count} and ISAC$_\text{LO}$.
}
\label{tab:count_results_extension}

\newcolumntype{C}[1]{>{\centering\arraybackslash}p{#1}}

\newlength{\countquantextensioncolwidth}
\setlength{\countquantextensioncolwidth}{0.06\textwidth} 
\resizebox{0.7\linewidth}{!}{
\begin{tabular}{
    l   
    *{5}{C{\countquantextensioncolwidth}}  
}
\toprule
\multirow{2}{*}{Method}
  & \multicolumn{5}{c}{IntraCompBench (Instance) (\(\uparrow\))} \\
\cmidrule(lr){2-6} 
  & \#2 & \#3 & \#4 & \#5 & Avg. \\
\midrule
SDXL \cite{podell2023sdxl}  & 90\% & 71\% & 49\% & 32\% & 61\% \\
+ CountGen {\tiny\color{gray}CVPR'25} \cite{binyamin2024count}  & \underline{97\%} & {83\%} & {52\%} & {44\%} & {69\%} \\ 
\rowcolor{drawioblue}
+ \textbf{ISAC$_\text{LO}$ (Ours)}  & {96\%} & \underline{89\%} & \underline{71\%} & \underline{47\%} & \underline{76\%} \\ 
\rowcolor{drawioblue}
+ CountGen + \textbf{ISAC$_\text{LO}$ (Ours)}  & \textbf{98\%} & \textbf{91\%} & \textbf{74\%} & \textbf{50\%} & \textbf{78\%} \\ 

\bottomrule
\end{tabular}
}
\end{table}

\begin{figure}[p]
    \centering
    \includegraphics[width=0.7\linewidth]{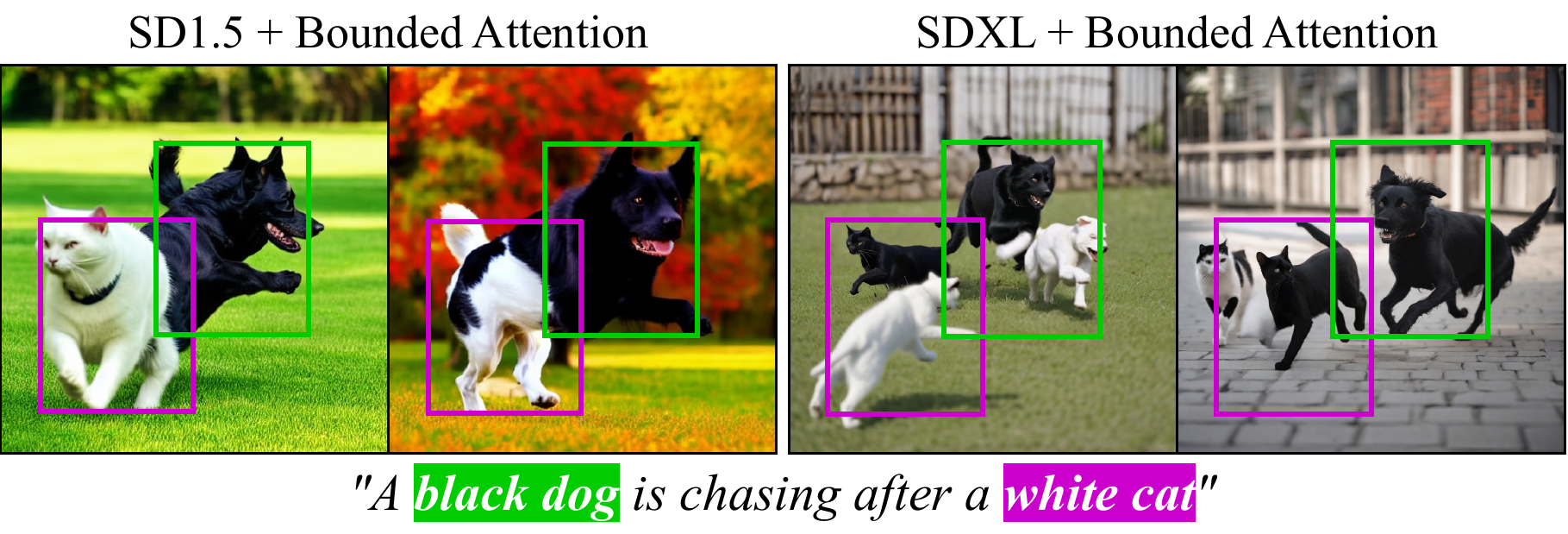}
    \caption{
    Semantic separation of layout guidance methods (\eg, Bounded Attention~\cite{dahary2024yourself}) do not ensure instance discrimination.
    }
    \label{fig:bounded_attention_separation}
\end{figure}

Among training‑free layout guidance methods~\cite{bar2023multidiffusion,shirakawa2024noisecollage,Xie_2023_ICCV,chen2023trainingfree,xiao2023rb,dahary2024yourself,lee2024groundit,lian2023llmgrounded,park2024rare}, Bounded Attention~\cite{dahary2024yourself} is the only approach that explicitly separates self‑ and cross‑attention across box-level masks to mitigate semantic mixing. Because it only partitions semantics by boxes, it cannot guarantee instance discrimination and often fails to count correctly in crowded scenes (see~\cref{fig:bounded_attention_separation}). 

\begin{figure}[p]
    \centering
    \includegraphics[width=0.7\linewidth]{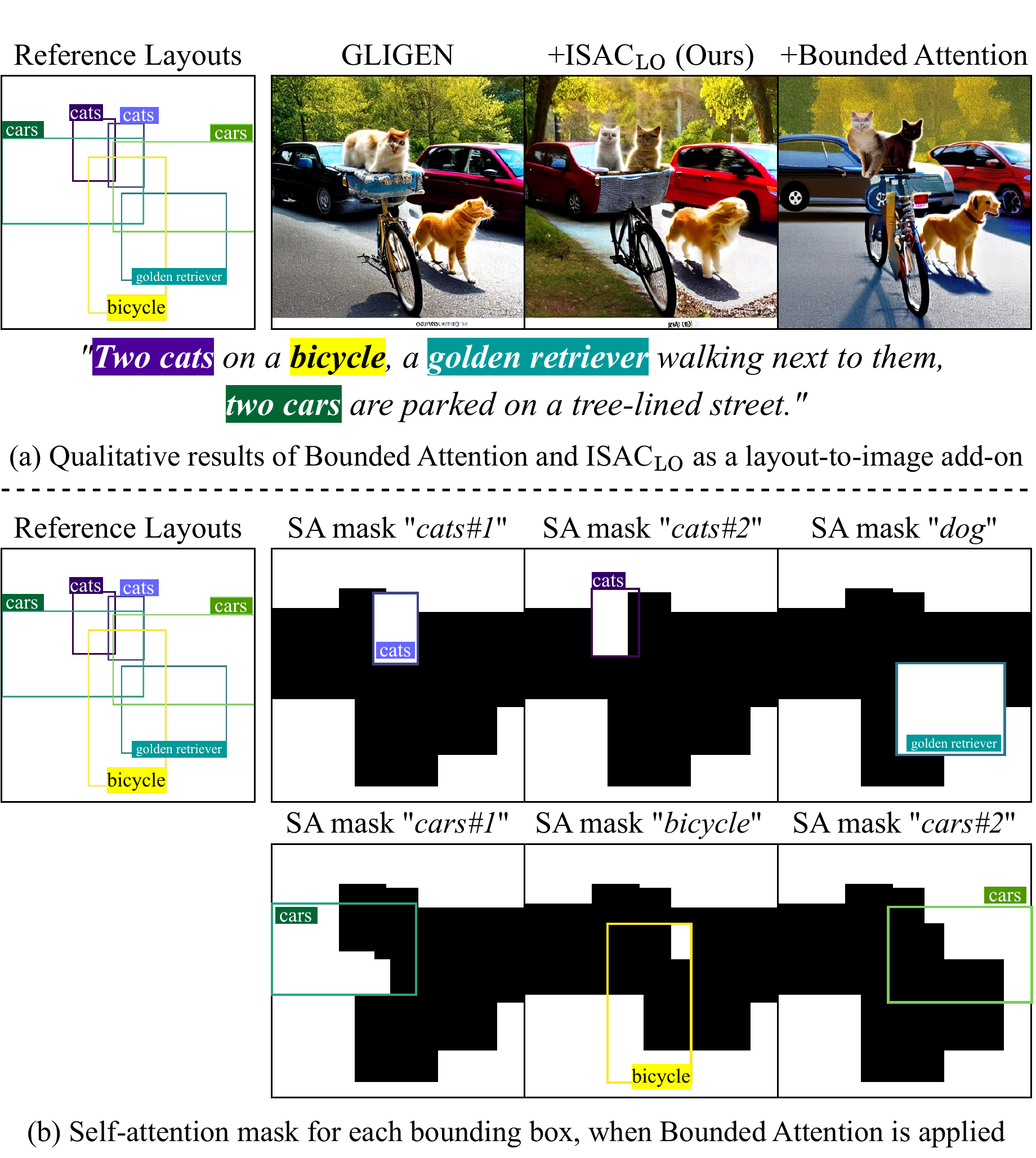}
    \caption{
    \textbf{Limitation of Bounded Attention masking.}
    Bounded Attention~\cite{dahary2024yourself} enforces exclusive pixel ownership among bounding boxes on self-attention maps. Each box can only attend to its owned pixels and the background.
    We adopt a ``smaller-box-first'' ownership rule to build self-attention masks. Each mask visualizes pixels attendable from the corresponding layout. As shown, exclusivity suppresses attention in shared regions and degrades control.
    }
    \label{fig:gligen_bounded_attention}
\end{figure}

Moreover, its mutual‑exclusivity assumption—each pixel belongs to at most one box —breaks under overlapping or tightly packed layouts. Thus, the resulting suppression of attention in shared regions degrades spatial control and semantic fidelity (\cref{fig:gligen_bounded_attention}). Since performance can also depend on the arbitrary ownership order under overlap, we exclude Bounded Attention from our layout comparisons.

\subsubsection{Fine-tuned models.}
Several methods ensure object‑level separation in attention maps by adapting the base model with external segmentation signals. TokenCompose~\cite{Wang2024TokenCompose} and CoMat~\cite{jiang2024comat} fine‑tune the UNet so that cross‑attention aligns with masks predicted by a pretrained model~\cite{ren2024grounded}. A key limitation is the reduced effective vocabulary compared with general‑purpose diffusion backbones, which restricts applicability. ISAC is complementary to these methods and can be applied at inference time without retraining (\cref{tab:isac-extension-fine-tuned}).

\clearpage
\section{Additional Discussions}
\label{sec:add-discussions}

\subsection{Multi-Instance Generation of Commercial Models}
\label{sec:commercial-models}

\begin{figure}[b]
  \centering
  \includegraphics[width=0.62\linewidth]{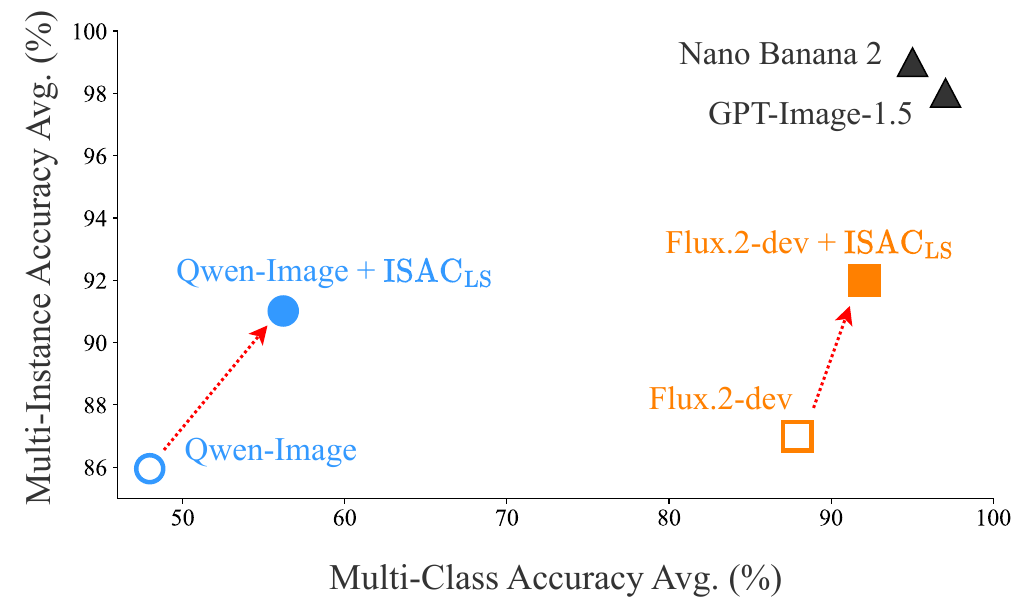}
  \caption{Two-dimensional comparison of average performance on IntraCompBench. Across both Qwen-Image \cite{wu2025qwenimagetechnicalreport} and Flux.2-dev \cite{flux-2-2025}, ISAC moves open-weight models toward the upper-right, showing consistent gains on both metrics and narrowing the gap to commercial models.}
  \label{fig:commercial-quant-plot}
\end{figure}

While commercial models such as GPT-Image-1.5 \cite{gptimage15} and Nano Banana 2 \cite{nanobanana2} establish a high upper bound for multi-instance compositionality, a notable performance disparity remains between these closed-source systems and base open-weight architectures like Qwen-Image \cite{wu2025qwenimagetechnicalreport} and Flux.2-dev \cite{flux-2-2025}. As demonstrated in \cref{tab:isac-commercial-extended} and \cref{fig:commercial-quant-plot}, applying the ISAC objective consistently shifts the performance of open-weight models toward this commercial upper bound. By utilizing explicit attention manipulation, ISAC effectively bridges the empirical gap between accessible open-weight models and resource-intensive closed-source systems without the need for architectural modifications or fine-tuning. 

\begin{table*}[ht]
\centering
\setlength{\tabcolsep}{3pt}       

\caption{Performance comparison between open-weight models equipped with ISAC and commercial models on IntraCompBench. ISAC can be seamlessly applied to state-of-the-art open-weight baselines, such as Qwen-Image \cite{wu2025qwenimagetechnicalreport} and Flux.2-dev \cite{flux-2-2025}. ISAC effectively narrows the performance gap with commercial counterparts \cite{gptimage15, nanobanana2}.}
\label{tab:isac-commercial-extended}

\newcolumntype{C}[1]{>{\centering\arraybackslash}p{#1}}

\newlength{\commercialcolwidth}
\setlength{\commercialcolwidth}{0.06\textwidth} 

\resizebox{\linewidth}{!}{
\begin{tabular}{
    l   
    *{5}{C{\commercialcolwidth}}  
    *{5}{C{\commercialcolwidth}}  
    c   
    c   
}
    \toprule
    \multirow{2}{*}{Method}
      & \multicolumn{5}{c}{Multi‑Class Accuracy (\(\uparrow\))}
      & \multicolumn{5}{c}{Multi‑Instance Accuracy (\(\uparrow\))}
      & \multicolumn{2}{c}{Efficiency (\(\downarrow\))} \\
    \cmidrule(lr){2-6} \cmidrule(lr){7-11} \cmidrule(lr){12-13}
      & \#2 & \#3 & \#4 & \#5 & Avg.
      & \#2 & \#3 & \#4 & \#5 & Avg.
      & Latency & VRAM \\
    \midrule

    \(\text{Qwen-Image}\) \cite{wu2025qwenimagetechnicalreport} & 91\% & 45\% & 33\% & 10\% & 48\% & 98\% & 92\% & 84\% & 70\% & 86\% & \textbf{140s} & \textbf{60.1GB} \\
    \rowcolor{drawioblue}
    + \(\textbf{ISAC}_\text{LS}\) \textbf{(Ours)}& \textbf{99\%} & \textbf{58\%} & \textbf{42\%} & \textbf{25\%} & \textbf{56\%} & \textbf{99\%} & \textbf{96\%} & \textbf{89\%} & \textbf{78\%} & \textbf{91\%} & 210s & 65.3GB \\

    \midrule
    \(\text{Flux.2-dev}\) \cite{flux-2-2025} & 97\% & 95\% & 84\% & 78\% & 88\% & \textbf{100\%} & 93\% & 81\% & 75\% & 87\% & \textbf{205s} & \textbf{74.2GB} \\
    \rowcolor{drawioblue}
    + \(\textbf{ISAC}_\text{LS}\) \textbf{(Ours)} & \textbf{99\%} & \textbf{98\%} & \textbf{89\%} & \textbf{83\%} & \textbf{92\%} & \textbf{100\%} & \textbf{98\%} & \textbf{88\%} & \textbf{81\%} & \textbf{92\%} & 305s & 79.8GB \\

    \midrule
    \rowcolor{gray!20}
    \(\text{GPT-Image-1.5}\) \cite{gptimage15} & 99\% & 99\% & 98\% & 95\% & 97\% & 100\% & 100\% & 99\% & 94\% & 98\% & \text{N/A} & \text{N/A} \\
    \rowcolor{gray!20}
    \(\text{Nano Banana 2}\) \cite{nanobanana2} & 99\% & 97\% & 93\% & 92\% & 95\% & 100\% & 100\% & 100\% & 95\% & 99\% & \text{N/A} & \text{N/A} \\
    
    \bottomrule
\end{tabular}%
}
\end{table*}

\begin{figure}[ht]
    \centering
    \includegraphics[width=0.6\linewidth]{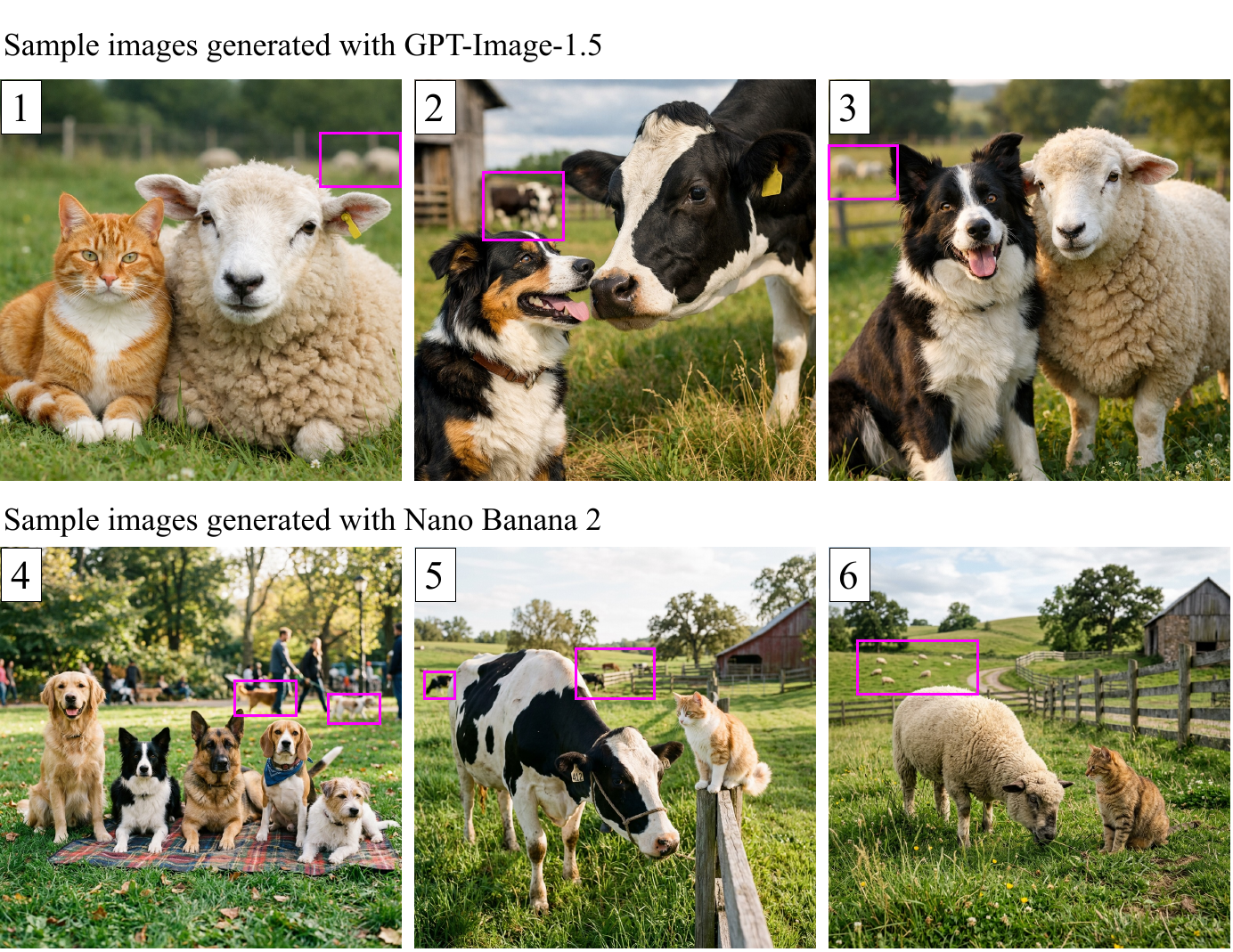}
    \caption{Multi-instance generation samples from GPT-Image-1.5 \cite{gptimage15} and Nano Banana 2 \cite{nanobanana2}. Used input prompts are: (1)\textit{``A photo of a cat and a sheep''}, (2)\textit{``A photo of a dog and a cow''}, (3)\textit{``A photo of a dog and a sheep''}, (4)\textit{``A photo of 5 dogs''}, (5)\textit{``A photo of a cat and a cow''}, (6)\textit{``A photo of a cat and a sheep''}.}
    \label{fig:commercial-samples}
\end{figure}

Despite their robust performance on standardized benchmarks, qualitative assessments indicate that multi-instance generation remains a nuanced challenge even for advanced commercial systems \cite{gptimage15,nanobanana2}; for example, they occasionally generate unwanted background instances when processing semantically adjacent classes (\cref{fig:commercial-samples}). These minor artifacts suggest that implicit text guidance does not perfectly resolve all boundary ambiguities, highlighting the continued scientific relevance of explicit structural interventions like ISAC.

\subsection{Detector-Agnostic Evaluation of ISAC}
\label{sec:detector_agnostic_evaluation}

In \cref{tab:detector_agnostic_evaluation}, we report a vision-language-model (VLM)-based evaluation of multi-instance generation, adopting the protocol of \cite{qiu2025self}. On the IntraCompBench suite, GPT-5.5~\cite{gpt-5.5} independently reproduces ISAC's gains in both magnitude and correlation. These results confirm that ISAC's improvements are not driven by detector-specific biases.

\begin{table}[hb!]
\centering
\setlength{\tabcolsep}{6pt}       

\caption{Detector and VLM evaluation on IntraCompBench.}
\label{tab:detector_agnostic_evaluation}

\resizebox{0.75\linewidth}{!}{
\begin{tabular}{
    l   
    c   
    c   
}
    \toprule
    Method &
    Detector Ensemble (\(\uparrow\)) &
    GPT-5.5~\cite{gpt-5.5} (\(\uparrow\)) \\
    \midrule 

    SD1.5~\cite{rombach2022high} & 8\% & 14\% \\
    \rowcolor{drawioblue}
    \(+\text{ISAC}_\text{LO}\) & \textbf{36\%} & \textbf{42\%} \\
    Pearson($r$) vs. Detectors & - & 0.73 \\
    \midrule

    SD3.5-M~\cite{esser2024scaling} & 25\% & 32\% \\
    \rowcolor{drawioblue}
    \(+\text{ISAC}_\text{LO}\) & \textbf{52\%} & \textbf{58\%} \\
    Pearson($r$) vs. Detectors & - & 0.75 \\
    
    \bottomrule
\end{tabular}%
}
\end{table}

\subsection{Effect on Perceptual Quality and Human Preference}
\label{sec:quality_analysis}

We analyze whether ISAC$_\text{LO}$'s structural interventions affect the perceptual quality of generated images. In \cref{tab:isac_image_quality}, we report \textit{Distribution Metrics} (Frechet Inception Distance (FID)~\cite{heusel2017gans} and CLIP~\cite{radford2021learning}‑based Maximum Mean Discrepancy (CMMD)~\cite{jayasumana2024rethinking}) alongside \textit{Human Preference Metrics} (PickScore~\cite{kirstain2023pick}, HPSv3~\cite{ma2025hpsv3}).

It is important to note that preference models like PickScore~\cite{kirstain2023pick} and HPSv3~\cite{ma2025hpsv3} evaluate generation quality holistically, considering both text-image alignment and aesthetic fidelity.
As shown in \cref{tab:isac_image_quality}, ISAC$_\text{LO}$ demonstrates a clear superiority in human preference, achieving a significantly higher PickScore win rate ($0.42 < 0.58$) and improved HPSv3 score ($6.72 \rightarrow 7.15$) compared to the baseline SDXL~\cite{podell2023sdxl}.
Given that ISAC$_\text{LO}$ dramatically improves instance accuracy (as seen in earlier evaluations), this preference gain confirms that our method does not trade off visual quality for controllability. Instead, it achieves a superior balance, generating images that are not only structurally accurate but also more aligned with human visual preferences.

Regarding distribution shifts, while FID~\cite{heusel2017gans} shows a slight increase, the CMMD~\cite{jayasumana2024rethinking} remains negligible ($0.0026$). Since CMMD~\cite{jayasumana2024rethinking} is more robust to outliers and better correlates with human perception than FID~\cite{jayasumana2024rethinking}, this result suggests ISAC keeps the native generative manifold without noticeable artifacts.

\begin{table}[t]
\centering
\setlength{\tabcolsep}{5pt}       
\caption{
\textbf{Performance-aesthetic trade-off analysis on SDXL.} 
We evaluate three aspects: Alignment (Class/Instance accuracy on IntraCompBench), Distribution (FID~\cite{heusel2017gans}/CMMD~\cite{jayasumana2024rethinking}), and Preference (PickScore~\cite{kirstain2023pick}/HPSv3~\cite{ma2025hpsv3}).
For distribution and preference metrics, we generate images using the T2I‑CompBench~\cite{huang2025t2ipp} Numeracy task.  FID and CMMD measure the distance between the generated distribution and the native SDXL~\cite{podell2023sdxl} baseline; lower values indicate less deviation from the original aesthetic manifold. Note that PickScore represents the win rate against the baseline.}

\label{tab:isac_image_quality}
\resizebox{\linewidth}{!}{
\begin{tabular}{l c c c c c c}
    \toprule
    \multirow{2}{*}{Method}
      & \multicolumn{2}{c}{Alignment Metrics}
      & \multicolumn{2}{c}{Distribution Metrics}
      & \multicolumn{2}{c}{Preference Metrics} \\
    \cmidrule(lr){2-3} \cmidrule(lr){4-5} \cmidrule(lr){6-7}
      & Class (\(\uparrow\)) & Instance (\(\uparrow\)) 
      & FID (\(\downarrow\)) & CMMD (\(\downarrow\))
      & PickScore (\(\uparrow\)) & HPSv3 (\(\uparrow\)) \\
        
    \midrule
    SDXL~\cite{podell2023sdxl}              & 7\%  & 61\%  & {\color{gray}0.0000}  & {\color{gray}0.0000} & 0.4208 & 6.7230 \\
    \rowcolor{drawioblue}
    + \textbf{ISAC$_\text{LO}$ (Ours)}                   & 34\% & 76\%  & 16.402  & 0.0026 & 0.5792 & 7.1463 \\
    \bottomrule
\end{tabular}
}
\end{table}

\subsection{Sensitivity to Internal Instance Count}

ISAC uses the parsed instance count $N$ to form $N$ self-attention clusters. To examine the effect of count mismatch, we perturb only the internal count used by ISAC to $N-1$ or $N+1$, while keeping the prompt and evaluation target unchanged. Relative to the default $N$ setting reported in \cref{tab:main-results}, multi-class accuracy drops by 22/16 percentage points for $N-1/N+1$ on SD1.5~\cite{rombach2022high} and by 25/18 percentage points on SD3.5-M~\cite{esser2024scaling}. This confirms that ISAC benefits from an accurate count prior. Using too few clusters forces distinct target instances to share a layout, weakening instance separation. On the other hand, using too many clusters fragments a single object into artificial sub-instances and disrupts semantic binding.

\subsection{Effect of Self-Attention Clustering Frequency}

\begin{figure}[ht]
    \centering
    \includegraphics[width=0.9\linewidth]{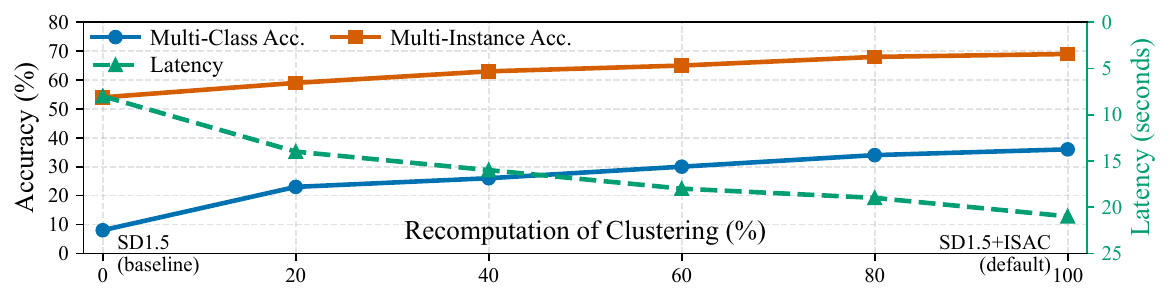}
    \caption{
    \textbf{Recomputation trade-off (SD1.5).}
    Per-step (100\%; right) maximizes accuracy; lower rate yields a smooth trade-off.
    }
    \label{fig:ablation_clustering_frequency}
\end{figure}

ISAC clusters self-attention maps at every denoising step. As this is time-consuming, \cref{fig:ablation_clustering_frequency} analyzes the trade-off between recomputation frequency and accuracy.
Together with the clustering algorithm comparison in \cref{tab:clustering_alternatives}, and the coordinate-augmentation analysis in \cref{fig:coord-concat}, this result shows that ISAC's clustering is robust to algorithmic choices but benefits from frequent recomputation to track evolving attention layouts.

\subsection{Choice of Loss Design} 
\Cref{tab:ablation-loss-type} evaluates alternative similarity metrics in place of MPO. Replacing MPO with MAE, KL, IoU or Dice consistently degrades performance, reducing multi-class accuracy from 36\% to at most 21\% and multi-instance accuracy from 69\% to at most 61\%. 
This gap indicates that MPO’s margin-based separation between class-specific masks is better suited to enforcing instance-aware semantic decoupling than generic overlap measures.

Top-$k$\% MPO alternatives, which average the largest $k$\% overlaps instead of taking the maximum, are smoother but less effective than MPO. This gap indicates that focusing on peak overlap reduction is an effective design choice.

\begin{table}[hb]
\centering
\caption{Analysis of alternative similarity metrics in \cref{eq:mpo}.}
\label{tab:ablation-loss-type}
\setlength{\tabcolsep}{6pt}  

\resizebox{0.65\linewidth}{!}{
\begin{tabular}{l l c c}
    \toprule
    \multirow{2}{*}{Method}  & \multirow{2}{*}{Metric}  & \multicolumn{2}{c}{IntraCompBench ($\uparrow$)} \\
    \cmidrule(lr){3-4}
            &         & Class & Instance \\
    \midrule
    SD1.5~\cite{rombach2022high} & N/A &  8\%  & 54\% \\
    + ISAC$_\text{LO}$  & MAE &  9\% {\color{ForestGreen}(+1\%)}  & 55\% {\color{ForestGreen}(+1\%)}  \\
    + ISAC$_\text{LO}$  & KL  & 16\% {\color{ForestGreen}(+8\%)}  & 60\% {\color{ForestGreen}(+6\%)}  \\
    + ISAC$_\text{LO}$  & IoU & 20\% {\color{ForestGreen}(+12\%)} & 61\% {\color{ForestGreen}(+7\%)}  \\
    + ISAC$_\text{LO}$  & Dice & 21\% {\color{ForestGreen}(+13\%)} & 61\% {\color{ForestGreen}(+7\%)} \\
    + ISAC$_\text{LO}$  & top-$10$\% MPO & 28\% {\color{ForestGreen}(+19\%)} & 63\% {\color{ForestGreen}(+9\%)} \\
    + ISAC$_\text{LO}$  & top-$5$\% MPO & 31\% {\color{ForestGreen}(+23\%)} & 65\% {\color{ForestGreen}(+11\%)} \\
    \rowcolor{drawioblue}
    + ISAC$_\text{LO}$  & MPO & \textbf{36\%} {\color{ForestGreen}(+28\%)} & \textbf{69\%} {\color{ForestGreen}(+15\%)} \\
    \bottomrule
\end{tabular}
}
\end{table}

\subsection{Performance Stability of ISAC}
Initial random seeds ($X_T$) significantly affect compositional generation~\cite{li2024all}. Thus, we report seed-wise stability on the T2I-CompBench Numeracy task in \cref{tab:add-quantitative-results-t2i} (IntraCompBench already averages 10 seeds/prompt). ISAC$_\text{LO}$ reduces standard deviation by $\sim$3-4$\times$, yielding qualitatively stable generations (\cref{fig:3cats_comparison}). This demonstrates that the sharp MPO loss does not induce generation instability.

\subsection{Integration with Fine-tuned Models}
To evaluate complementarity, we apply ISAC to two externally supervised fine-tuned methods: TokenCompose~\cite{Wang2024TokenCompose}, using Grounded SAM masks~\cite{ren2024grounded} for cross-attention alignment, and IterComp~\cite{zhang2024itercomp}, employing preference-guided refinement with RPG~\cite{yang2024mastering} and InstanceDiffusion~\cite{wang2024instancediffusion}. ISAC further improves multi-class and multi-instance accuracy (\cref{tab:isac-extension-fine-tuned}). Incidentally, TokenCompose requires \texttt{float32} weights, increasing latency and VRAM over standard SD1.4~\cite{rombach2022high}.

\begin{table*}[ht]
\centering
\setlength{\tabcolsep}{3pt}       

\caption{IntraCompBench evaluation of ISAC$_\text{LO}$ on fine-tuned models \cite{Wang2024TokenCompose, zhang2024itercomp}}
\label{tab:isac-extension-fine-tuned}

\newcolumntype{C}[1]{>{\centering\arraybackslash}p{#1}}

\newlength{\finetunecolwidth}
\setlength{\finetunecolwidth}{0.055\textwidth} 

\resizebox{\linewidth}{!}{
\begin{tabular}{
    l   
    *{5}{C{\finetunecolwidth}}  
    *{5}{C{\finetunecolwidth}}  
    c   
    c   
}
    \toprule
    \multirow{2}{*}{Method}
      & \multicolumn{5}{c}{Multi‑Class Accuracy (\(\uparrow\))}
      & \multicolumn{5}{c}{Multi‑Instance Accuracy (\(\uparrow\))}
      & \multicolumn{2}{c}{Efficiency (\(\downarrow\))} \\
    \cmidrule(lr){2-6} \cmidrule(lr){7-11} \cmidrule(lr){12-13}
      & \#2 & \#3 & \#4 & \#5 & Avg.
      & \#2 & \#3 & \#4 & \#5 & Avg.
      & Latency & VRAM \\
    \midrule
    \(\text{TokenCompose}_\text{SD1.4}\) \cite{Wang2024TokenCompose}     & 27\% &  4\% &  1\% &  0\% &  8\% & 77\% & 65\% & 41\% & 13\% & 49\% & \textbf{12s} & \textbf{8.5GB} \\
    \rowcolor{drawioblue}
    + \textbf{ISAC$_\text{LO}$ (Ours)}                                                & \textbf{62\%} & \textbf{36\%} & \textbf{28\%} & \textbf{17\%} & \textbf{36\%} & \textbf{84\%} & \textbf{80\%} & \textbf{60\%} & \textbf{30\%} & \textbf{63\%} & 33s & 17.9GB \\
    
    \midrule
    \(\text{IterComp}_\text{SDXL}\) \cite{zhang2024itercomp}     & 11\% &  5\% &  4\% &  0\% &  5\% & 95\% & 73\% & 64\% & 37\% & 67\% & \textbf{49s} & \textbf{11.8GB} \\
    \rowcolor{drawioblue}
    + \textbf{ISAC$_\text{LO}$ (Ours)}                                        & \textbf{46\%} & \textbf{28\%} & \textbf{26\%} & \textbf{21\%} & \textbf{30\%} & \textbf{99\%} & \textbf{93\%} & \textbf{85\%} & \textbf{55\%} & \textbf{83\%} & 100s & 29.9GB \\
    
    \bottomrule
    \end{tabular}%
}
\end{table*}

\subsection{Limitation}

\begin{figure}[hb]
    \centering
    \includegraphics[width=0.7\linewidth]{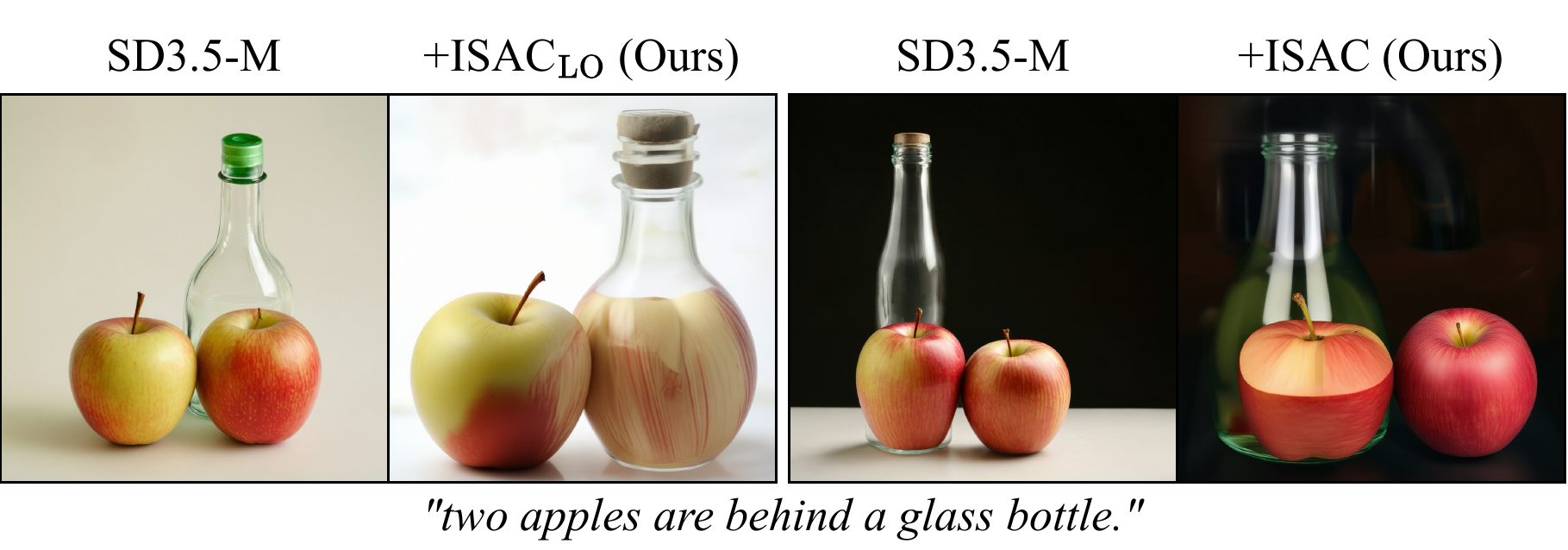}
    \caption{Limitation of ISAC.}
    \label{fig:limitation}
\end{figure}

ISAC lacks explicit 3D understanding, which can fail for prompts requiring depth ordering through transparent materials, as shown in Fig.~\ref{fig:limitation}. Future work will explore 3D- or physics-aware extensions.

\clearpage
\section{Future Works}
\label{sec:future_work}

\begin{figure}[ht]
    \centering
    \includegraphics[width=0.65\linewidth]{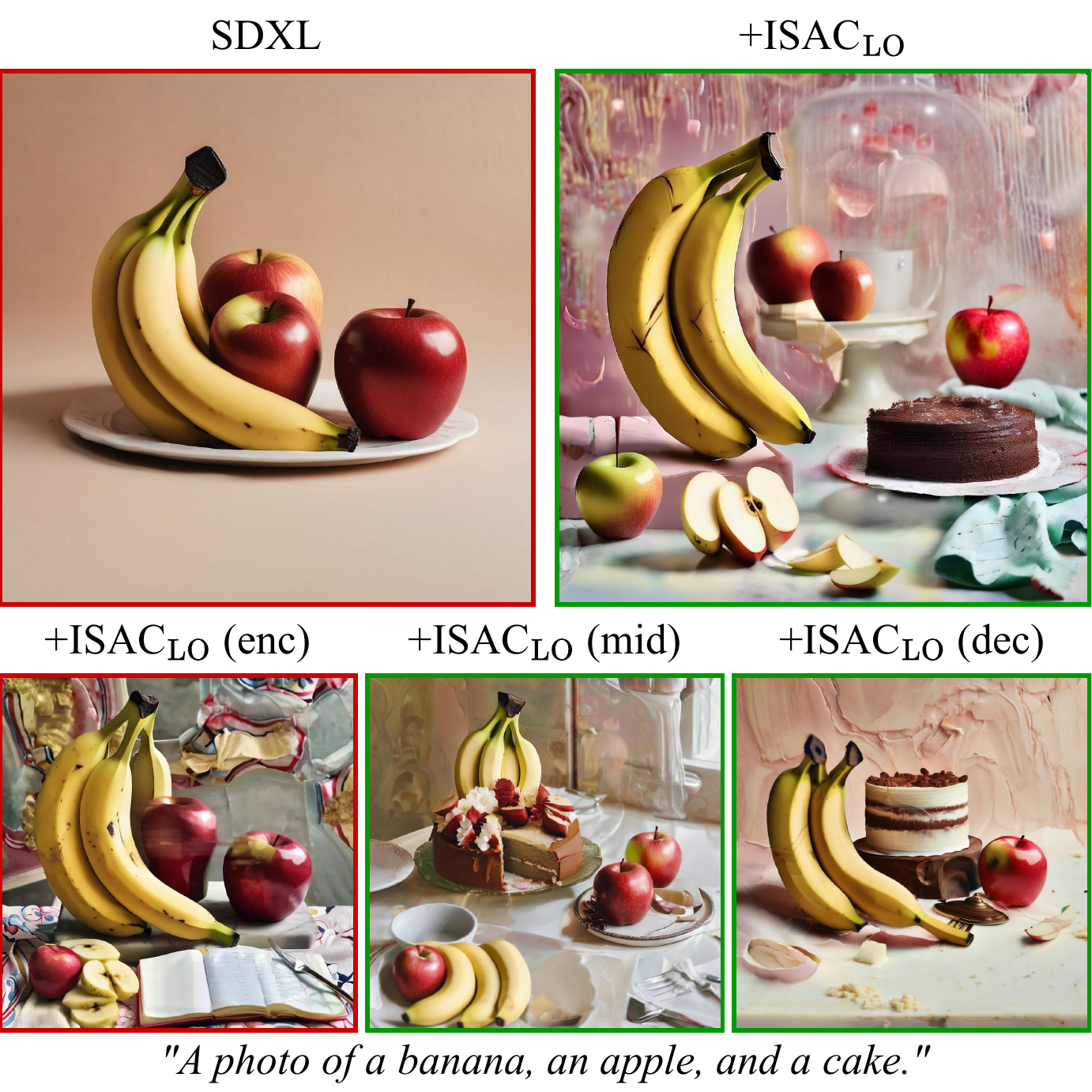}
    \caption{ISAC$_\text{LO}$ applied to different subsets of SDXL~\cite{podell2023sdxl} attention layers.}
    \label{fig:future_work}
\end{figure}

Most of ISAC's additional computational cost comes from applying backpropagation through \emph{all} attention layers in the diffusion model. A natural way to reduce this overhead is to identify informative attention layers and apply backpropagation only to those. Ahn \etal \cite{ahn2024self} explore such informative layers and empirically find that specific attention layers in the mid-block of the U-Net~\cite{ronneberger2015u} are particularly important for guidance. \cref{fig:future_work} and \cref{tab:future_work} show that ISAC exhibits a similar trend, suggesting that layer-wise selection is a promising direction for improving ISAC's efficiency.

\begin{table}[ht]
\centering
\setlength{\tabcolsep}{5pt}       
\caption{Complexity analysis of SDXL~\cite{podell2023sdxl} and ISAC$_\text{LO}$ on various layers.}
\label{tab:future_work}
\begin{tabular}{lcc}
    \toprule
    Method & Latency (s) (\(\downarrow\)) & VRAM (GB) (\(\downarrow\)) \\
    \midrule
    SDXL~\cite{podell2023sdxl} w/o attention & \textbf{10.7}  & \textbf{18.0} \\
    + ISAC$_\text{LO}$           & 111.4 & 30.7 \\
    + ISAC$_\text{LO}$ (enc)     & 60.9  & \underline{21.2} \\
    + ISAC$_\text{LO}$ (mid)     & \underline{31.7}  & \textbf{18.0} \\
    + ISAC$_\text{LO}$ (dec)     & 60.4  & 25.0 \\
    \bottomrule
\end{tabular}
\end{table}

\clearpage
\section{Additional Diffusion Dynamics Visualization}
\label{sec:dynamics_analysis}

\begin{figure*}[p]
    \centering
    \includegraphics[width=\textwidth]{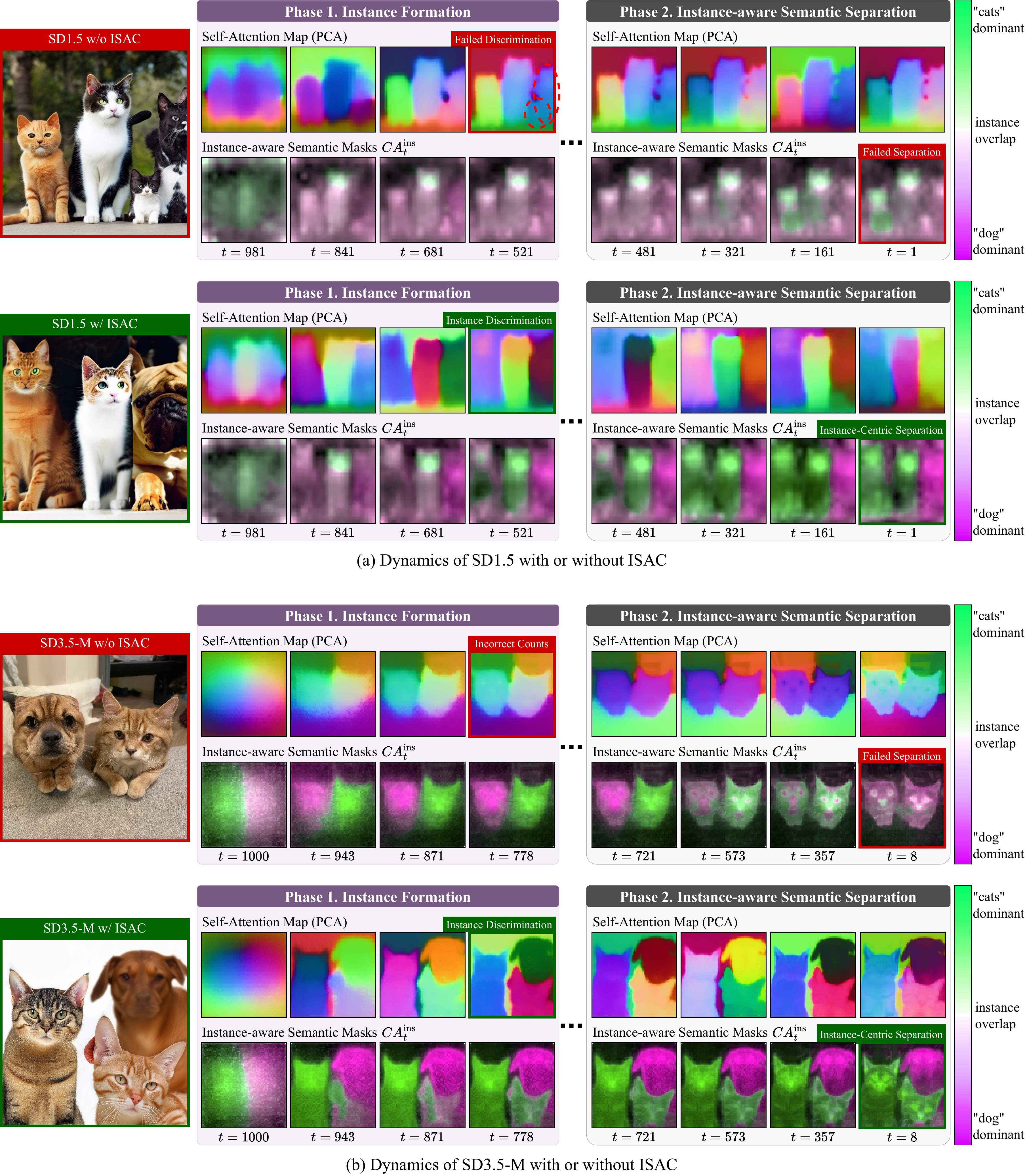}
    \caption{
    \textbf{Diffusion dynamics visualization with or without ISAC$_\text{LO}$.}
    This is an extended version of \cref{fig:method-overview} to SD1.5~\cite{rombach2022high} and SD3.5-M~\cite{esser2024scaling}.
    }
    \label{fig:more_method_dynamics}
\end{figure*}

In \cref{fig:more_method_dynamics}, we show the extended visualization of \cref{fig:method-overview} to popular diffusion models \cite{rombach2022high,esser2024scaling} and ISAC$_\text{LO}$'s dynamics on SD3.5-M~\cite{esser2024scaling}. The two-step instance formation and instance-centric semantic separation achieves highly reliable multi-object generation.

\section{Additional Quantitative Results}
\label{sec:add-quantitative-results}

We provide full quantitative results on IntraCompBench metrics in \Cref{tab:add-quantitative-results}. The results are obtained by applying ISAC$_\text{LO}$ to various models, including SD1.4, SD1.5, SD2.1 \cite{rombach2022high}, SDXL \cite{podell2023sdxl}, SD3.5-M, SD3.5-L \cite{esser2024scaling}, PixArt-\(\alpha\) \cite{chen2023pixartalpha}, PixArt-\(\Sigma\) \cite{chen2024pixart} and Flux.1-dev \cite{flux2024}.

Also in \cref{tab:add-quantitative-results-t2i}, we provide extended quantitative results of ISAC$_\text{LO}$ on IntraCompBench and public counting benchmark T2I-CompBench~\cite{huang2025t2ipp}, where prompts specify exact counts. It is worth note that SD1.5~\cite{rombach2022high} with ISAC$_\text{LO}$ surpasses  baseline performance of plain SDXL~\cite{podell2023sdxl} and ISAC$_\text{LO}$ reduces performance variance from initial seeds (see~\cref{sec:add-discussions} for details).

\begin{table*}[p]
\centering
\setlength{\tabcolsep}{3pt}        

\caption{Additional quantitative results of latent optimization methods on IntraCompBench.}
\label{tab:add-quantitative-results}

\newcolumntype{C}[1]{>{\centering\arraybackslash}p{#1}}

\newlength{\myaddquantcolwidth}
\setlength{\myaddquantcolwidth}{0.06\textwidth} 

\resizebox{\linewidth}{!}{
\begin{tabular}{
    l   
    *{5}{C{\myaddquantcolwidth}}  
    *{5}{C{\myaddquantcolwidth}}  
    c   
    c   
}
    \toprule
    \multirow{2}{*}{Method}
      & \multicolumn{5}{c}{Multi‑Class Accuracy (\(\uparrow\))}
      & \multicolumn{5}{c}{Multi‑Instance Accuracy (\(\uparrow\))}
      & \multicolumn{2}{c}{Efficiency (\(\downarrow\))} \\
    \cmidrule(lr){2-6} \cmidrule(lr){7-11} \cmidrule(lr){12-13}
      & \#2 & \#3 & \#4 & \#5 & Avg.
      & \#2 & \#3 & \#4 & \#5 & Avg.
      & Latency & VRAM \\
    \midrule
    SD1.4 \cite{rombach2022high}                & 30\% &  2\% &  1\% &  0\% &  8\% & 94\% & 74\% & 28\% & 22\% & 55\% &  \textbf{8s} &  \textbf{4.9GB} \\
    + A\&E {\tiny\color{gray}SIGGRAPH'23} \cite{chefer2023attendandexcite}     & 50\% &  9\% &  8\% &  2\% & 17\% & 97\% & 79\% & 26\% & 23\% & 56\% & 17s &  9.2GB \\
    + SynGen {\tiny\color{gray}NeurIPS'23} \cite{rassin2024linguistic}        & 54\% & 11\% &  6\% &  2\% & 18\% & 90\% & 69\% & 25\% & 19\% & 51\% & 19s &  9.3GB \\
    + InitNO {\tiny\color{gray}CVPR'24} \cite{guo2024initno}               & 58\% & 10\% &  7\% &  4\% & 20\% & 94\% & 79\% & 31\% & 20\% & 56\% & 20s &  9.6GB \\
    + TEBOpt {\tiny\color{gray}NeurIPS'24} \cite{Chen_2024_TEBOpt}            & 55\% & 13\% &  8\% &  2\% & 19\% & 91\% & 73\% & 31\% & 20\% & 54\% & 17s &  9.3GB \\
    \rowcolor{drawioblue}
    + \textbf{ISAC$_\text{LO}$ (Ours)}                       & \textbf{66\%} & \textbf{34\%} & \textbf{29\%} & \textbf{16\%} & \textbf{36\%} &\textbf{100\%} & \textbf{90\%} & \textbf{51\%} & \textbf{40\%} & \textbf{70\%} & 21s &  9.7GB \\

    \midrule
    SD1.5 \cite{rombach2022high}                & 28\% & 2\%  & 1\%  & 0\%  & 8\%  & 88\% & 65\% & 36\% & 26\% & 54\% & \textbf{12s} & \textbf{4.4 GB} \\
    + A\&E {\tiny\color{gray}SIGGRAPH'23} \cite{chefer2023attendandexcite}     & 48\% & 10\% & 5\%  & 2\%  & 16\% & 91\% & 68\% & 34\% & 24\% & 54\% & 24s &  9.1 GB \\
    + SynGen {\tiny\color{gray}NeurIPS'23} \cite{rassin2024linguistic}        & 50\% & 9\%  & 4\%  & 2\%  & 16\% & 84\% & 61\% & 38\% & 22\% & 51\% & 27s &  9.2 GB \\
    + InitNO {\tiny\color{gray}CVPR'24} \cite{guo2024initno}               & 55\% & 12\% & 7\%  & 5\%  & 20\% & 90\% & 68\% & 40\% & 29\% & 57\% & 29s &  9.5 GB \\
    + Self-Cross {\tiny\color{gray}CVPR'25} \cite{qiu2025self}             & 48\% &  8\% & 4\% &  2\%  & 15\% & 89\% & 67\% & 38\% & 27\% & 55\% & 21s & 10 GB  \\
    + TEBOpt {\tiny\color{gray}NeurIPS'24} \cite{Chen_2024_TEBOpt}            & 52\% & 11\% & 8\%  & 3\%  & 18\% & 87\% & 65\% & 36\% & 27\% & 54\% & 25s &  9.2 GB \\
    \rowcolor{drawioblue}
    + \textbf{ISAC$_\text{LO}$ (Ours)}                       & \textbf{65\%} & \textbf{31\%} & \textbf{29\%} & \textbf{18\%} & \textbf{36\%} & \textbf{95\%} & \textbf{82\%} & \textbf{56\%} & \textbf{44\%} & \textbf{69\%} & 30s & 9.6 GB \\

    \midrule
    SD2.1 \cite{rombach2022high}                & 31\% &  6\% &  3\% &  0\% & 10\% & 91\% & 74\% & 41\% & 28\% & 58\% & \textbf{13s} &  \textbf{4.8 GB} \\
    + A\&E {\tiny\color{gray}SIGGRAPH'23} \cite{chefer2023attendandexcite}     & 53\% & 12\% &  4\% &  1\% & 18\% & 94\% & 79\% & 39\% & 29\% & 60\% & 26s &  9.3 GB \\
    + SynGen {\tiny\color{gray}NeurIPS'23} \cite{rassin2024linguistic}        & 55\% & 10\% &  7\% &  3\% & 19\% & 87\% & 69\% & 38\% & 25\% & 55\% & 29s &  9.4 GB \\
    + InitNO {\tiny\color{gray}CVPR'24} \cite{guo2024initno}               & 59\% & 13\% & 11\% &  5\% & 22\% & 91\% & 79\% & 44\% & 26\% & 60\% & 31s &  9.7 GB \\
    + TEBOpt {\tiny\color{gray}NeurIPS'24} \cite{Chen_2024_TEBOpt}            & 56\% & 14\% &  7\% &  6\% & 21\% & 88\% & 75\% & 44\% & 27\% & 58\% & 27s &  9.4 GB \\
    \rowcolor{drawioblue}
    + \textbf{ISAC$_\text{LO}$ (Ours)}                       & \textbf{67\%} & \textbf{35\%} & \textbf{34\%} & \textbf{20\%} & \textbf{39\%} & \textbf{98\%} & \textbf{88\%} & \textbf{64\%} & \textbf{42\%} & \textbf{73\%} & 32s &  9.8 GB \\

    \midrule
    SDXL \cite{podell2023sdxl}                  & 20\% &  4\% &  3\% &  0\% &  7\% & 90\% & 71\% & 49\% & 32\% & 61\% & \textbf{48s} & \textbf{12.8 GB} \\
    \rowcolor{drawioblue}
    + \textbf{ISAC$_\text{LO}$ (Ours)}                       & \textbf{57\%} & \textbf{32\%} & \textbf{29\%} & \textbf{17\%} & \textbf{34\%} & \textbf{96\%} & \textbf{89\%} & \textbf{71\%} & \textbf{47\%} & \textbf{76\%} & 101s & 29.8 GB \\

    \midrule
    PixArt-\(\alpha\) \cite{chen2023pixartalpha}& 27\% &  3\% &  1\% &  0\% &  8\% & 99\% & 93\% & 33\% & 15\% & 60\% & \textbf{17s} & \textbf{19.9 GB} \\
    \rowcolor{drawioblue}
    + \textbf{ISAC$_\text{LO}$ (Ours)}                       & \textbf{63\%} & \textbf{30\%} & \textbf{29\%} & \textbf{21\%} & \textbf{36\%} &\textbf{100\%} &\textbf{100\%} & \textbf{56\%} & \textbf{31\%} & \textbf{72\%} & 40s & 53.7 GB \\

    \midrule
    PixArt-\(\Sigma\) \cite{chen2024pixart}     & 39\% &  8\% &  0\% &  0\% & 12\% & 98\% & 98\% & 30\% & 16\% & 60\% & \textbf{18s} & \textbf{19.9 GB} \\
    \rowcolor{drawioblue}
    + \textbf{ISAC$_\text{LO}$ (Ours)}                       & \textbf{78\%} & \textbf{39\%} & \textbf{31\%} & \textbf{20\%} & \textbf{42\%} & \textbf{100\%} & \textbf{100\%} & \textbf{48\%} & \textbf{31\%} & \textbf{70\%} & 41s & 53.8 GB \\

    \midrule
    SD3.5-M \cite{esser2024scaling}             & 62\% & 23\% & 12\% & 3\%  & 25\% & 84\% & 71\% & 51\% & 51\% & 64\% & \textbf{40s} &\textbf{22.9 GB} \\
    + A\&E {\tiny\color{gray}SIGGRAPH'23} \cite{chefer2023attendandexcite}     & 65\% & 29\% &  16\% &  5\% & 28\% & 86\% & 72\% & 52\% & 50\% & 65\% & 124s  &  73.8 GB \\
    + SynGen {\tiny\color{gray}NeurIPS'23} \cite{rassin2024linguistic}        & 66\% & 28\% &  15\% &  6\% & 28\% & 82\% & 68\% & 50\% & 48\% & 62\% & 131s &  74.3 GB  \\
    + InitNO {\tiny\color{gray}CVPR'24} \cite{guo2024initno}               & 77\% & 31\% &  17\% &  7\% & 33\% & 84\% & 73\%  & 52\% & 49\% & 65\% & 138s &  74.6 GB  \\
    + Self-Cross {\tiny\color{gray}CVPR'25} \cite{qiu2025self}             & 78\% & 38\% &  19\% &  3\% & 34\% & 86\% & 72\% & 51\% & 50\% & 65\% & 147s & 76.4 GB \\
    + TEBOpt {\tiny\color{gray}NeurIPS'24} \cite{Chen_2024_TEBOpt}            & 78\% & 31\% &  19\% &  8\% & 34\% & 85\% & 71\% & 52\% & 52\% & 65\% & 139s &  74.5 GB \\
    \rowcolor{drawioblue}
    + \textbf{ISAC$_\text{LO}$ (Ours)}                       & \textbf{98\%} & \textbf{51\%} & \textbf{40\%} & \textbf{20\%} & \textbf{52\%} & \textbf{98\%} & \textbf{91\%} & \textbf{72\%} & \textbf{69\%} & \textbf{83\%} & 140s & 74.8 GB \\

    \bottomrule
  \end{tabular}%
}
\end{table*}

\begin{table*}[p]
\centering
\setlength{\tabcolsep}{3pt}       

\caption{Quantitative results on T2I-CompBench~\cite{huang2025t2ipp} Numeracy and IntraCompBench multi-instance tasks. Performance stability across initial random seeds is measured on T2I-CompBench, with 30 seeds. \textbf{Bold} shows the best and \underline{underline} shows the second best performance.}
\label{tab:add-quantitative-results-t2i}

\newcolumntype{C}[1]{>{\centering\arraybackslash}p{#1}}

\newlength{\ttoinumcolwidth}
\setlength{\ttoinumcolwidth}{0.08\textwidth} 

\resizebox{\linewidth}{!}{
\begin{tabular}{
l   
c   
c   
*{5}{C{\ttoinumcolwidth}}  
    }
        \toprule
        \multirow{2}{*}{Method}
          &  \multirow{2}{*}{\# Parameters}
          & ~~~T2I-CompBench~~~
          & \multicolumn{5}{c}{IntraCompBench (Multi‑Instance) (\(\uparrow\))} \\
        \cmidrule(lr){3-3} \cmidrule(lr){4-8}
          & & Numeracy (\(\uparrow\))
          & \#2 & \#3 & \#4 & \#5 & Avg. \\
        \midrule
        SD1.5 \cite{rombach2022high}   & 0.8B & 46.5\% $\pm$ 5.1\% & 88\% & 65\% & 36\% & 26\% & 54\% \\
        \rowcolor{drawioblue}
        + \textbf{ISAC$_\text{LO}$ (Ours)}          & 0.8B & \underline{55.0\%} $\pm$ 1.4\% & \underline{95\%} & \underline{82\%} & \underline{56\%} & \underline{44\%} & \underline{69\%} \\
        \midrule
        SDXL \cite{podell2023sdxl}     & 2.6B & 51.1\% $\pm$ 4.8\% & 90\% & 71\% & 49\% & 32\% & 61\% \\
        \rowcolor{drawioblue}
        + \textbf{ISAC$_\text{LO}$ (Ours)}          & 2.6B & \textbf{64.0\%} $\pm$ 1.2\% & \textbf{96\%} & \textbf{89\%} & \textbf{71\%} & \textbf{47\%} & \textbf{76\%} \\
        \bottomrule
    \end{tabular}%
}
\end{table*}

\section{Additional Qualitative Results}
\label{appendix_qualitative}

\subsubsection{Generality beyond simple prompts.}
\begin{figure*}[p]
    \centering
    \includegraphics[width=\textwidth]{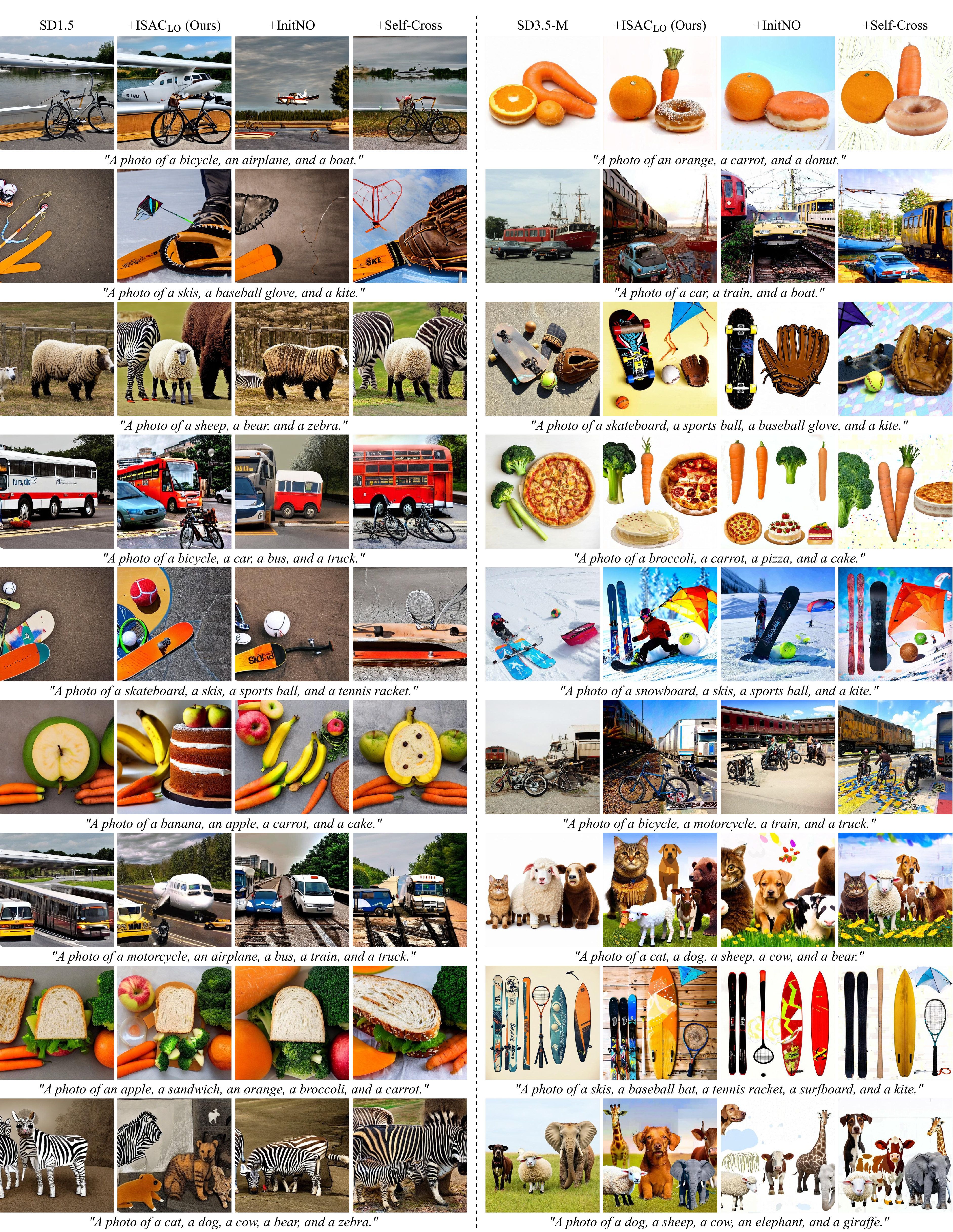}
    \caption{
    Qualitative comparisons of attention control methods, InitNO~\cite{guo2024initno}, Self-Cross~\cite{qiu2025self} and ISAC$_\text{LO}$, in 3 to 5 intra-category generation. All prompts are drawn from IntraCompBench.
    }
    \label{fig:qual_more_intra}
\end{figure*}

\begin{figure*}[p]
    \centering
    \includegraphics[width=\textwidth]{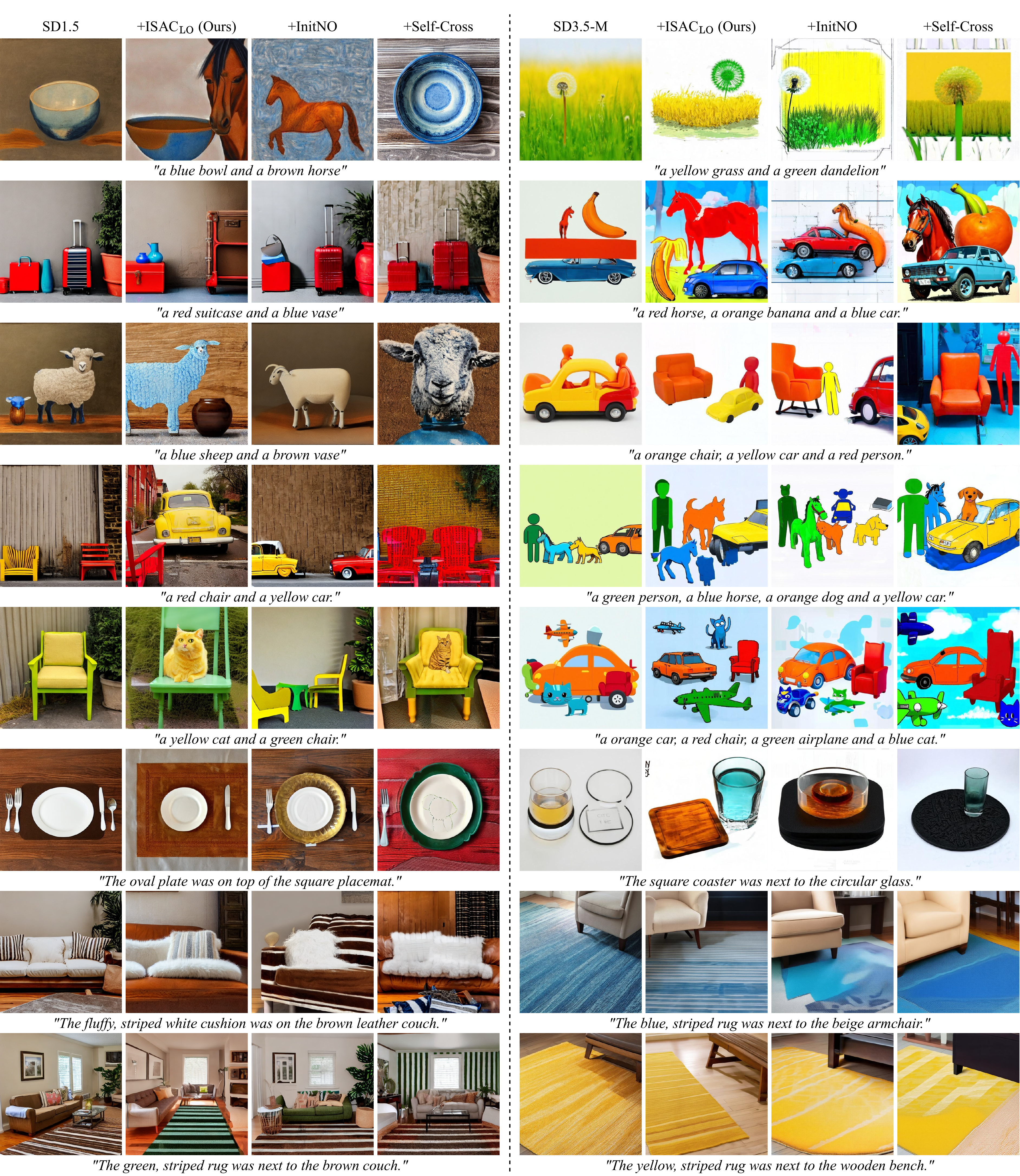}
    \caption{
    Qualitative comparisons of attention control methods, InitNO~\cite{guo2024initno}, Self-Cross~\cite{qiu2025self} and ISAC$_\text{LO}$, in complex scene generation. It requires the model to correctly bind attributes such as color, texture, spatial, and shape. All prompts are drawn from HRS-Bench~\cite{bakr2023hrs} and T2I-CompBench~\cite{huang2025t2ipp}.
    }
    \label{fig:qual_more_complex}
\end{figure*}

\begin{figure*}[p]
    \centering
    \includegraphics[width=\textwidth]{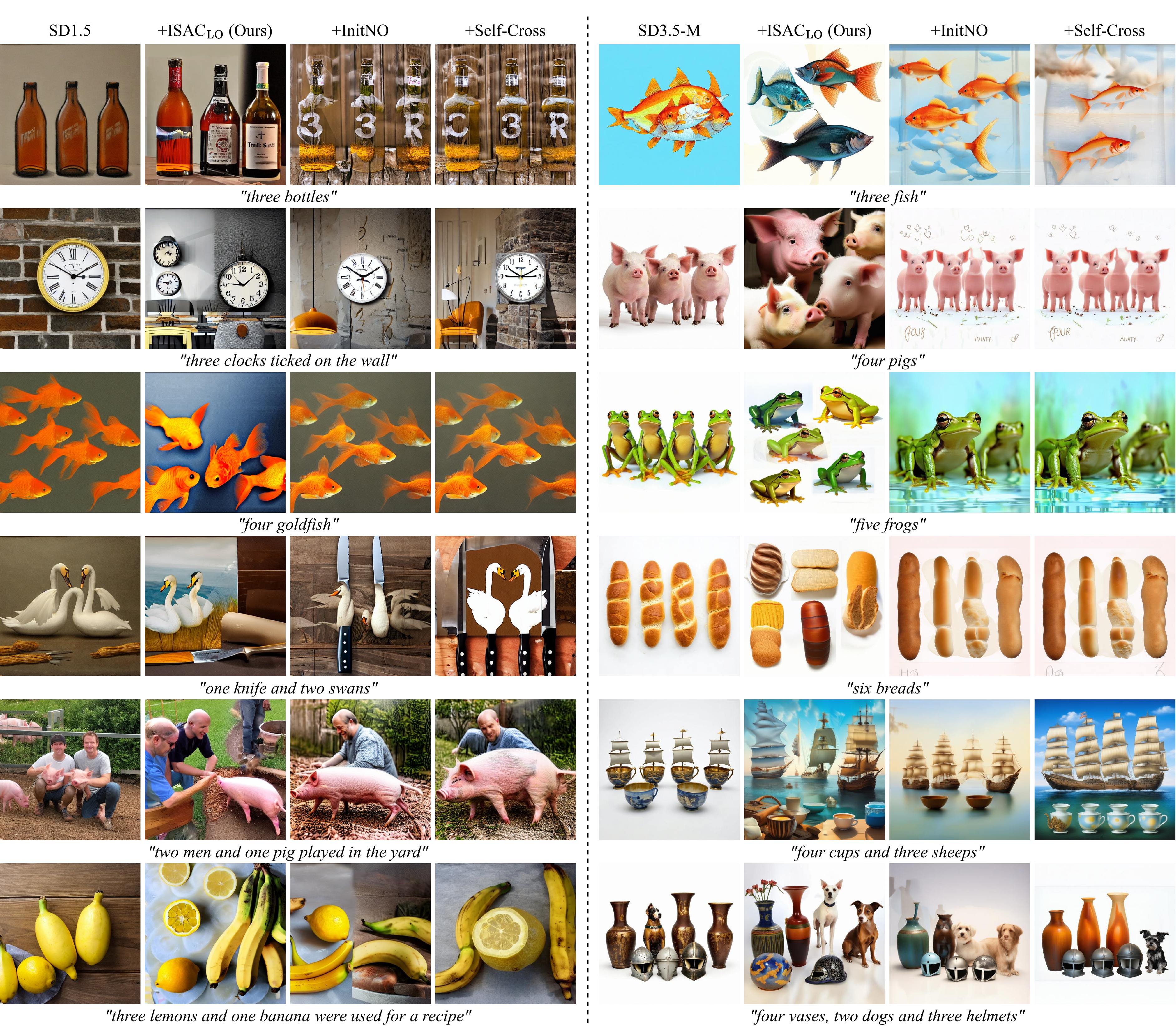}
    \caption{
    Qualitative comparisons of attention control methods, InitNO~\cite{guo2024initno}, Self-Cross~\cite{qiu2025self} and ISAC$_\text{LO}$, in general multi-instance scenarios. It requires the model to correctly generate multiple instances from one or more classes. All prompts are drawn from T2I-CompBench~\cite{huang2025t2ipp} Numeracy task.
    }
    \label{fig:qual_more_numeracy}
\end{figure*}

\begin{figure*}[p]
    \centering
    \includegraphics[width=\textwidth]{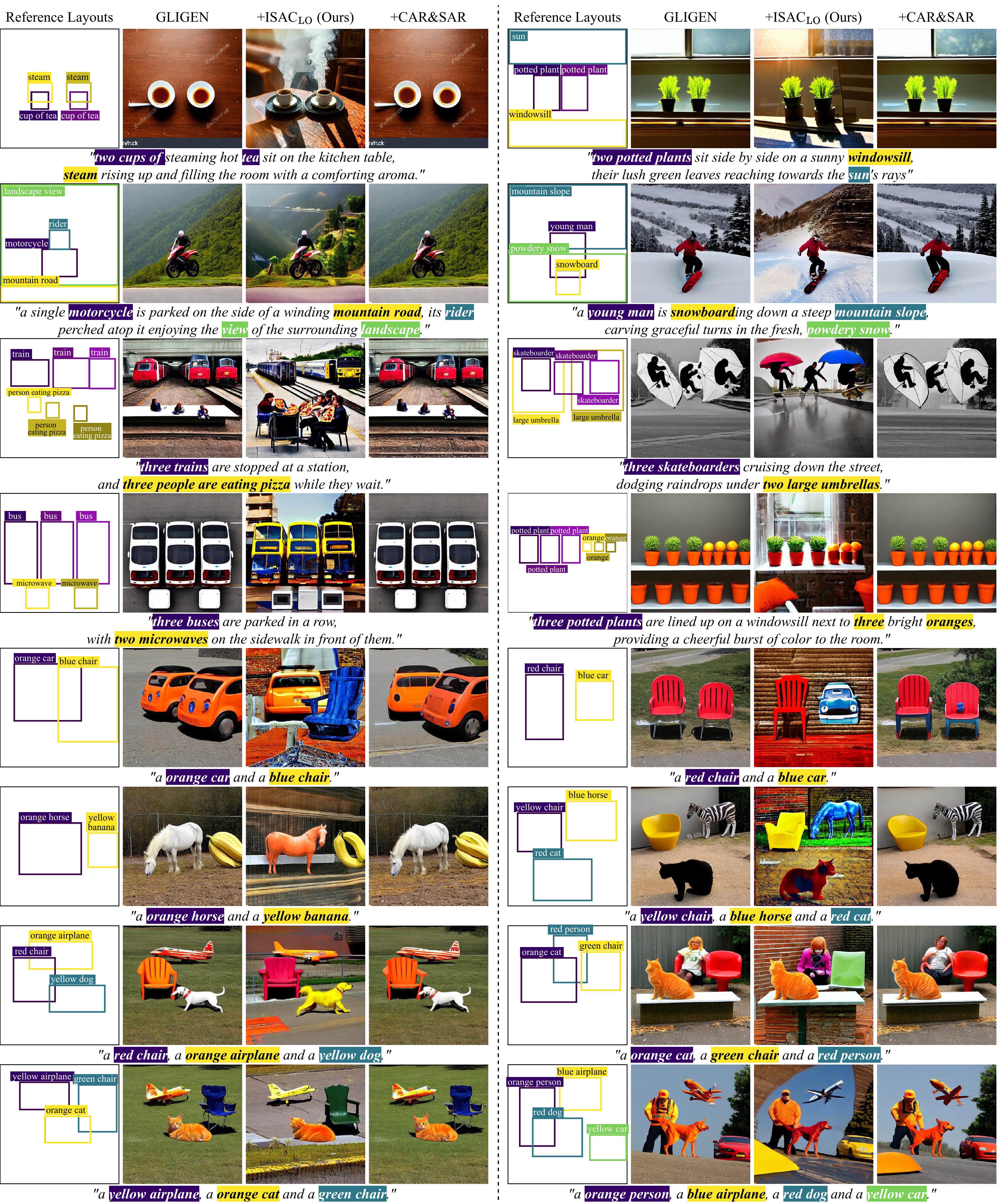}
    \caption{
    Qualitative comparisons of ISAC$_\text{LO}$ with layout guidance method, Attention Refocusing (CAR\&SAR)~\cite{phung2024grounded} on top of finetuned layout model GLIGEN~\cite{li2023gligen}. All prompts are drawn from HRS-Bench~\cite{huang2025t2ipp}.
    }
    \label{fig:qual_more_layout}
\end{figure*}

\Cref{fig:qual_more_intra,fig:qual_more_complex,fig:qual_more_numeracy,fig:qual_more_layout} show additional qualitative comparisons of ISAC$_\text{LO}$ with other attention control methods,  InitNO~\cite{guo2024initno}, Self-Cross~\cite{qiu2025self} and Attention-Refocusing~\cite{phung2024grounded}. The results show that ISAC$_\text{LO}$ not only improves drawing multiple instances, but also improves assigning correct attributes to each instance. Beyond simple color attributes, ISAC$_\text{LO}$ is also effective to shape, texture, positioning and the combination of them. This highlights ISAC's broader applicability.

\subsubsection{Robust multi-instance generation across seeds.} 
\begin{figure*}[t]
    \centering
    \begin{subfigure}[b]{0.8\linewidth}
        \centering
        \includegraphics[width=\linewidth]{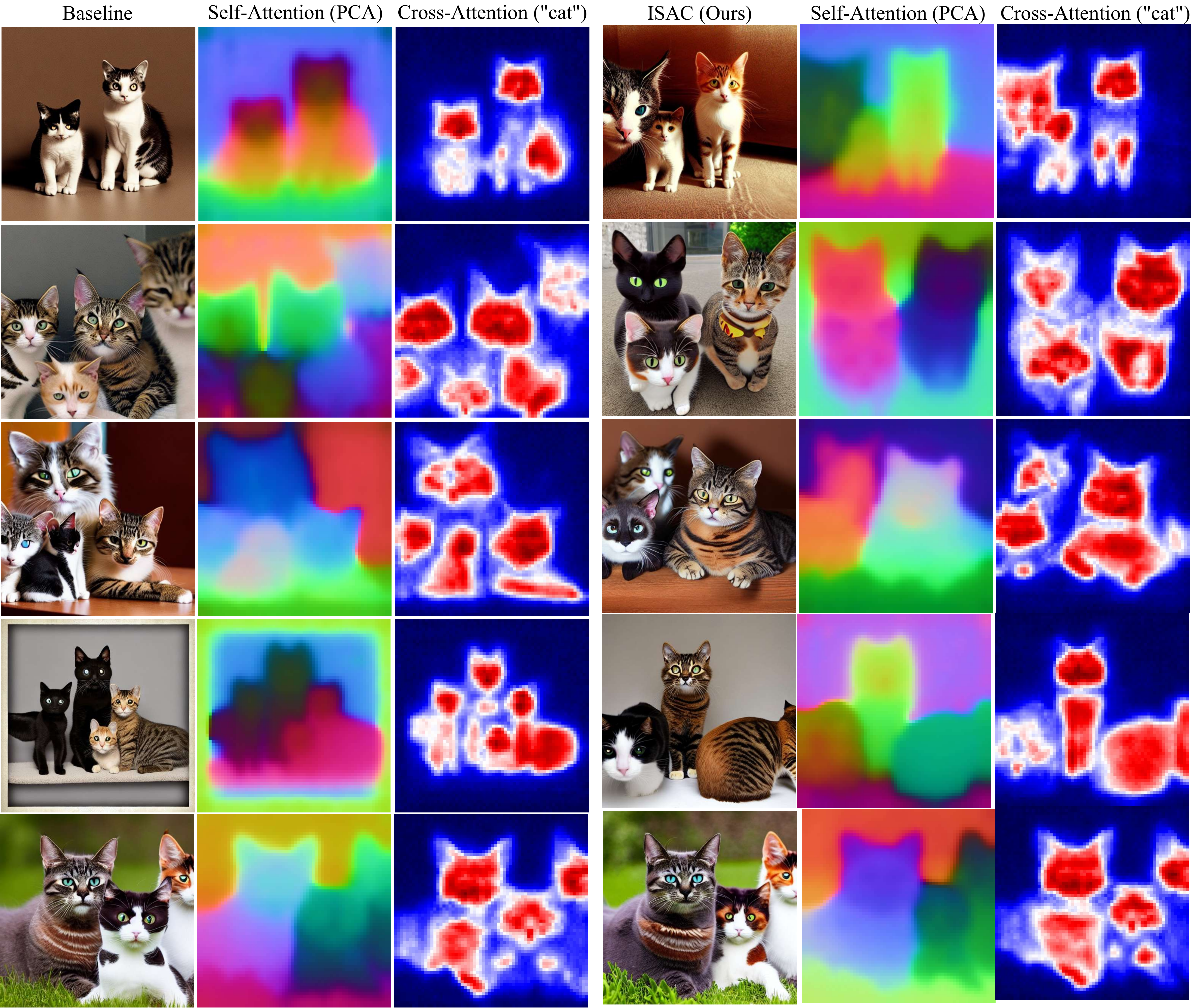}
        \caption{Result from Stable Diffusion v1.5 \cite{rombach2022high}.}
        \label{fig:3cats_sd15}
    \end{subfigure}
    \hfill
    \begin{subfigure}[b]{0.8\linewidth}
        \centering
        \includegraphics[width=\linewidth]{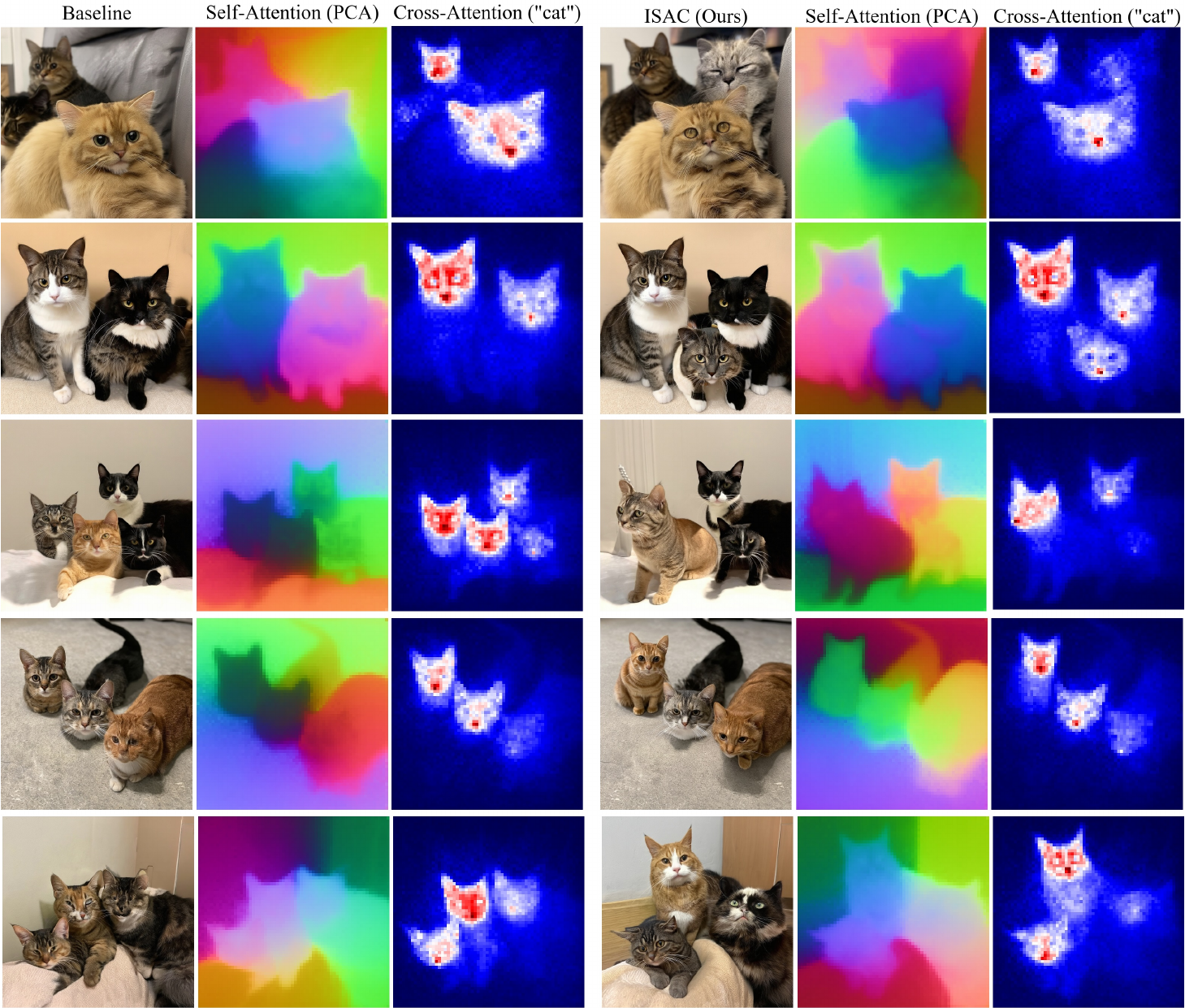}
        \caption{Result from Stable Diffusion v3.5-M \cite{esser2024scaling}.}
        \label{fig:3cats_sd35m}
    \end{subfigure}
    \caption{Qualitative results across multiple seeds for the prompt \textit{``three cats''}, which involves multiple instances of the same class. Our method (right) consistently generates images with the correct instance count, sharper object boundaries, and improved separation compared to the baseline (left).}
    \label{fig:3cats_comparison}
\end{figure*}

We provide qualitative results when generating 5 images with a fixed prompt, \textit{``three cats''}, on SD1.5 \cite{rombach2022high} and SD3.5-M \cite{esser2024scaling}. When the baseline output already contains the correct number of instances, ISAC$_\text{LO}$ minimally alters the result. However, when the baseline produces too few or too many instances, ISAC$_\text{LO}$ effectively corrects the output (see \cref{fig:3cats_sd15,fig:3cats_sd35m}). These results show that ISAC reliably helps correct instance counts across diverse seeds for a fixed prompt, while leaving correct samples mostly unchanged.

\end{document}